\pdfoutput=1

\documentclass[11pt]{article}

\usepackage{acl}

\usepackage{times}
\usepackage{latexsym}
\usepackage{hyperref}
\usepackage[utf8]{inputenc}
\usepackage{url}
\usepackage{booktabs}
\usepackage{subcaption}

\usepackage{multirow}
\usepackage{amsmath}
\usepackage{xcolor}
\usepackage{newfloat}
\usepackage{listings}
\usepackage{graphicx}
\usepackage{comment}
\usepackage{algorithm}
\usepackage{algorithmic}
\usepackage[T1]{fontenc}

\usepackage[utf8]{inputenc}

\usepackage{microtype}

\usepackage{inconsolata}

\usepackage[table]{xcolor}
\usepackage{tabularx}
\usepackage{booktabs}
\usepackage{geometry}

\usepackage{pifont}

\newcommand{\eg}{\textit{e.g.}}
\newcommand\book{\raisebox{-3pt}{\includegraphics[width=1.2em]{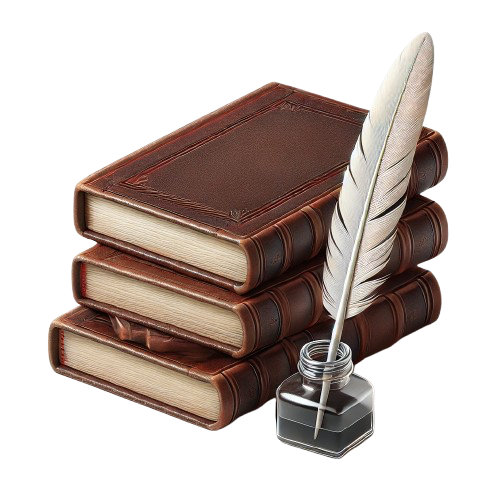}}}
\newcommand{\Principles}{\textsc{Principles}}

\title{\book \Principles: Synthetic Strategy Memory for Proactive Dialogue Agents}

\author{
    Namyoung Kim~~~
    Kai Tzu-iunn Ong~~~
    Yeonjun Hwang~~~
    Minseok Kang~~~ \\
    \textbf{Iiseo Jihn}~~~
    \textbf{Gayoung Kim}~~~
    \textbf{Minju Kim}~~~
    \textbf{Jinyoung Yeo}\thanks{Corresponding author} \\\\
    Department of Artificial Intelligence, Yonsei University \\
    \texttt{\{namyoung.kim, jinyeo\}@yonsei.ac.kr}\\
  }
\usepackage{indentfirst}
\begin{document}
\maketitle
\definecolor{green}{RGB}{36, 214, 36}
\definecolor{red}{RGB}{235, 30, 30}
\definecolor{lightredshade}{HTML}{dea9a9}
\definecolor{lightgreenshade}{HTML}{bce3bd}
\definecolor{lightblueshade}{HTML}{cacbe8}
\definecolor{MyDarkBlue}{rgb}{0,0.08,1}
\definecolor{MyDarkGreen}{rgb}{0.02,0.6,0.02}
\definecolor{MyDarkRed}{rgb}{0.8,0.02,0.02}
\definecolor{MyDarkOrange}{rgb}{0.40,0.2,0.02}
\definecolor{MyPurple}{RGB}{111,0,255}
\definecolor{MyRed}{rgb}{1.0,0.0,0.0}
\definecolor{MyGold}{rgb}{0.75,0.6,0.12}
\definecolor{MyDarkgray}{rgb}{0.66, 0.66, 0.66}

\definecolor{MyYellow}{rgb}{254, 246, 170}
\definecolor{MyBlue}{rgb}{170, 217, 251}
\definecolor{LuneBlue}{rgb}{0.11, 0.11, 0.43}

\newcommand{\greencheck}{\textcolor{green}{\ding{51}}}
\newcommand{\redcross}{\textcolor{red}{\ding{55}}}

\newcommand{\jy}[1]{\textcolor{MyDarkBlue}{[Jinyoung: #1]}}
\newcommand{\sy}[1]{\textcolor{MyDarkBlue}{[Shunyu: #1]}}
\newcommand{\hj}[1]{\textcolor{MyPurple}{[HJ: #1]}}
\newcommand{\sj}[1]{\textcolor{LuneBlue}{[SJ: #1]}}

\newcommand{\mcal}[1]{{\cal{#1}}}
\newcommand{\calA}{\mbox{${\cal A}$}}
\newcommand{\calB}{\mbox{${\cal B}$}}
\newcommand{\calC}{\mbox{${\cal C}$}}
\newcommand{\calD}{\mbox{${\cal D}$}}
\newcommand{\calE}{\mbox{${\cal E}$}}
\newcommand{\calF}{\mbox{${\cal F}$}}
\newcommand{\calG}{\mbox{${\cal G}$}}
\newcommand{\calH}{\mbox{${\cal H}$}}
\newcommand{\calI}{\mbox{${\cal I}$}}
\newcommand{\calJ}{\mbox{${\cal J}$}}
\newcommand{\calK}{\mbox{${\cal K}$}}
\newcommand{\calL}{\mbox{${\cal L}$}}
\newcommand{\calM}{\mbox{${\cal M}$}}
\newcommand{\calN}{\mbox{${\cal N}$}}
\newcommand{\calO}{\mbox{${\cal O}$}}
\newcommand{\calP}{\mbox{${\cal P}$}}
\newcommand{\calQ}{\mbox{${\cal Q}$}}
\newcommand{\calR}{\mbox{${\cal R}$}}
\newcommand{\calS}{\mbox{${\cal S}$}}
\newcommand{\calT}{\mbox{${\cal T}$}}
\newcommand{\calU}{\mbox{${\cal U}$}}
\newcommand{\calV}{\mbox{${\cal V}$}}
\newcommand{\calW}{\mbox{${\cal W}$}}
\newcommand{\calX}{\mbox{${\cal X}$}}
\newcommand{\calY}{\mbox{${\cal Y}$}}
\newcommand{\calZ}{\mbox{${\cal Z}$}}


\newcommand{\todocccc}[2]{{\textcolor{#1}{[[#2]]}}}
\newcommand{\todogreen}[1]{\todocccc{green}{[[#1]]}}
\newcommand{\haejuu}[1]{\todogreen{haejuu: #1}}
\newcommand{\yeo}[1]{\textcolor{purple}{#1}}
\newcommand{\dragon}[1]{\textcolor{brown}{#1}}
\newcommand{\bwoo}[1]{\textcolor{olive}{#1}}
\newcommand{\kyle}[1]{\textcolor{blue}{#1}} 
\newcommand{\cheris}[1]{\textcolor{teal}{#1}}
\newcommand{\chatgpt}[1]{\textcolor{gray}{#1}}
\newcommand{\minus}[1]{\textcolor{red}{#1}}
\newcommand{\plus}[1]{\textcolor{ForestGreen}{#1}}
\newcommand{\person}{{$\mathbb{P}$}}
\newcommand{\thought}{$\mathbb{Z}_{\mathbb{P}}$}
\newcommand{\ty}[1]{\textcolor{darkgreen}{[TY: #1]}}

\newcommand{\se}{{\it SE}}%
\newcommand{\etal}{{\it et al.}}%
\newcommand{\etc}{{\it etc}}%

\newcommand{\worldmodel}{$\mathcal{W}_{\theta}$}
\newcommand{\ours}{\textsc{Think-and-Execute}\xspace}
\newcommand{\coffeegym}{\textsc{Coffee-Gym}\xspace}
\newcommand{\cf}{\textsc{Coffee}\xspace}
\newcommand{\coffeeeval}{\textsc{CoffeeEval}\xspace}
\newcommand{\editeval}{\textsc{CoffeeEval}\xspace}
\newcommand{\editevalbf}{\textbf{\textsc{CoffeeEval}}\xspace}
\newcommand{\coffeewemoji}{\coffee\xspace\cf}
\newcommand{\coffeewemojibf}{\coffee\xspace\textbf{\cf}}
\newcommand{\cfwemoji}{\coffee\xspace\cf}
\newcommand{\cp}{\textsc{CoffeePots}\xspace}
\newcommand{\coffeepots}{\textsc{CoffeePots}}
\newcommand{\argmin}{\operatornamewithlimits{argmin}}
\newcommand{\argmax}{\operatornamewithlimits{argmax}}

\newcommand{\blueText}[1]{\textcolor{blue}{#1}}
\newcommand{\greenText}[1]{\textcolor{darkgreen}{#1}}

\lstdefinestyle{python}{
    language=Python,
    basicstyle=\fontsize{8}{10}\ttfamily,
    keywordstyle=\color{blue},
    commentstyle=\color{gray},
    stringstyle=\color{black},
    showstringspaces=false,
    breaklines=true,
    breakindent=0pt,
    breakatwhitespace=false,
    escapeinside={(*@}{@*)}
}

\lstdefinestyle{cpp}{
    language=C++,
    basicstyle=\fontsize{8}{10}\ttfamily,
    keywordstyle=\color{blue},
    commentstyle=\color{gray},
    stringstyle=\color{green},
    showstringspaces=false,
    breaklines=true,
    breakindent=0pt,
    breakatwhitespace=false,
    escapeinside={(*@}{@*)}
}

\lstdefinestyle{plain}{
    basicstyle=\fontsize{8}{10}\ttfamily,
    keywordstyle=\color{blue},
    commentstyle=\color{gray},
    stringstyle=\color{green},
    showstringspaces=false,
    breaklines=true,
    breakatwhitespace=false,
    breakindent=0pt,
    escapeinside={(*@}{@*)}
}

\lstdefinestyle{python2}{
    language=Python,
    basicstyle=\fontsize{8}{10}\ttfamily,
    keywordstyle=\color{blue},
    commentstyle=\color{gray},
    stringstyle=\color{green},
    showstringspaces=false,
    breakatwhitespace=false,
    breaklines=true,
    breakindent=0pt,
    escapeinside={(*@}{@*)}
}

\lstdefinestyle{cpp2}{
    language=C++,
    basicstyle=\fontsize{8}{10}\ttfamily,
    keywordstyle=\color{blue},
    commentstyle=\color{gray},
    stringstyle=\color{green},
    showstringspaces=false,
    breaklines=true,
    breakindent=0pt,
    breakatwhitespace=false,
    escapeinside={(*@}{@*)}
}

\lstdefinestyle{sql}{
    language=SQL,
    basicstyle=\fontsize{8}{10}\ttfamily,
    keywordstyle=\color{blue},
    commentstyle=\color{green},
    stringstyle=\color{black},
    showstringspaces=false,
    breakatwhitespace=false,
    breaklines=true,
    breakindent=0pt,
    escapeinside={(*@}{@*)}
}

\lstdefinestyle{prompt}{
    language=Python,
    basicstyle=\fontsize{8}{10}\ttfamily,
    keywordstyle=\color{blue},
    commentstyle=\color{gray},
    stringstyle=\color{cppgreen},
    showstringspaces=false,
    breaklines=true,
    backgroundcolor=\color{bgcolor},
    keepspaces=true, 
    breakindent=0pt,
    breakatwhitespace=false,
    showspaces=false,   
    escapeinside={(*@}{@*)}
}
\lstdefinestyle{text}{
    basicstyle=\fontsize{8}{10}\ttfamily,
    showstringspaces=false,
    breaklines=true,
    backgroundcolor=\color{bgcolor},
    breakatwhitespace=false,
    breakindent=0pt,
    keepspaces=true,
    showspaces=false,   
    escapeinside={(*@}{@*)}
}

\newcommand{\inserthl}[1]{\sethlcolor{promptinsert}\hl{#1}}
\newcommand{\comphl}[1]{\sethlcolor{compcolor}\hl{#1}}
\newcommand{\codehl}[1]{\sethlcolor{codehlcolor}\hl{#1}}
\newcommand{\codehlerr}[1]{\sethlcolor{codehlcolor2}\hl{#1}}


\begin{abstract}

Dialogue agents based on large language models (LLMs) have shown promising performance in proactive dialogue, which requires effective strategy planning.
However, existing approaches to strategy planning for proactive dialogue face several limitations: limited strategy coverage, preference bias in planning, and reliance on costly additional training.
To address these, we propose \textbf{\book \Principles}: a synthetic strategy memory for proactive dialogue agents. 
\Principles{} is derived through offline self-play simulations and serves as reusable knowledge that guides strategy planning during inference, eliminating the need for additional training and data annotation.
We evaluate \Principles{} in both emotional support and persuasion domains, demonstrating consistent improvements over strong baselines.
Furthermore, \Principles{} maintains its robustness across extended and more diverse evaluation settings. See our project page at \url{https://huggingface.co/spaces/kimnamssya/Principles}.
\end{abstract}

\section{Introduction}

\begin{figure}[t!]
    \centering
    \includegraphics[width=1\linewidth]{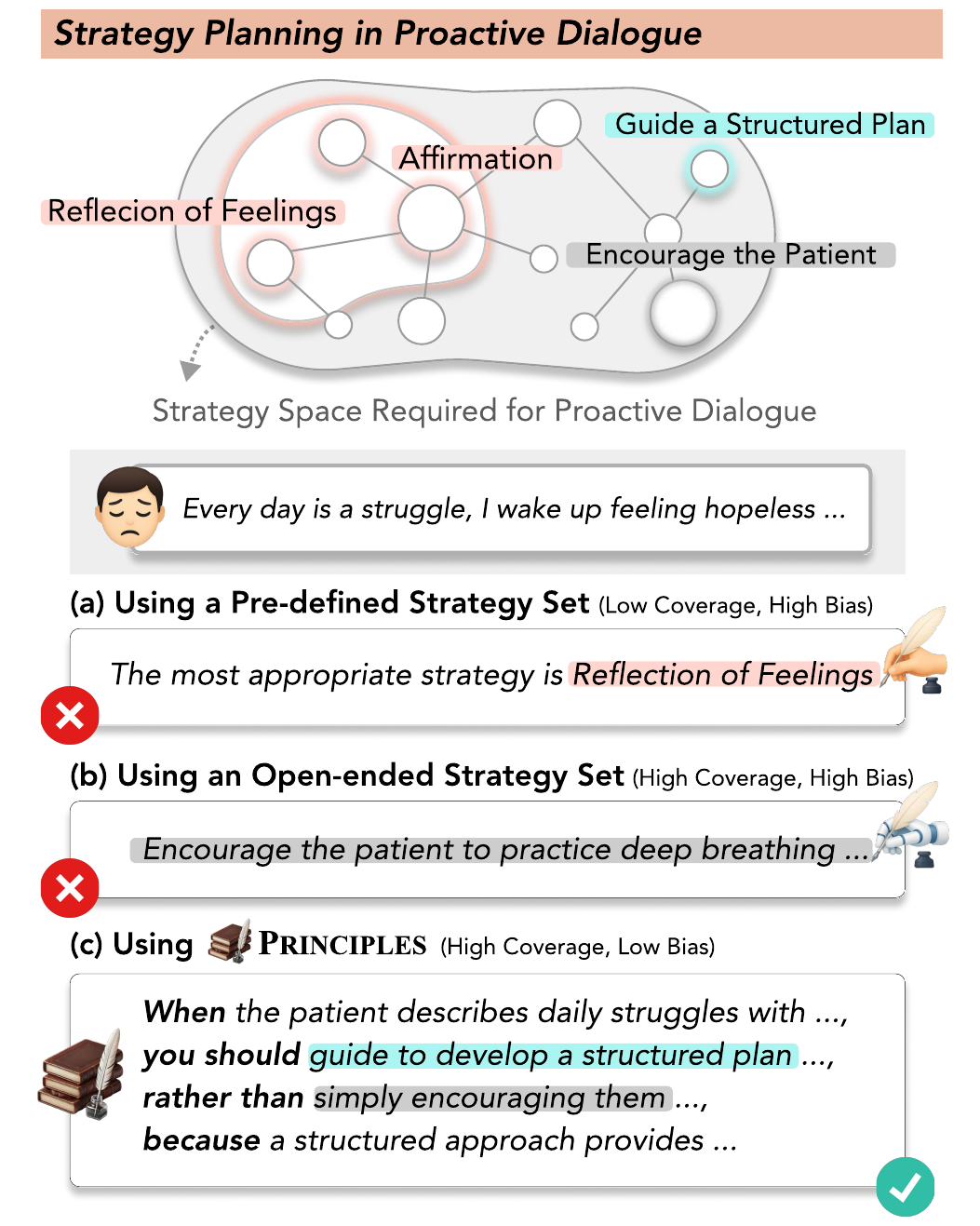}
    \caption{Empirical examples of strategy planning in proactive dialogue. (a) Pre-defined strategies fail due to limited coverage, (b) Open-ended strategies improve coverage but suffer from preference bias, (c) Our approach based on \Principles{} resolves both limited coverage and preference bias, leading to optimal outcomes.}
    \label{fig:figure_challenge}
\end{figure}

Recent advances in large language models (LLMs) have substantially improved the performance of dialogue agents~\cite{xu2022beyond, wang-etal-2023-divide, ong2025towards}. In this context, one primary research focus is improving dialogue agents' ability to plan strategies for achieving goals in proactive dialogue \citep{deng2023survey}, such as emotional support \citep{liu2021towards} and persuasion \citep{wang2019persuasion}.
A common approach is to employ an external planner to guide dialogue agents in selecting appropriate strategies based on the current context. To this end, \citet{zhang-etal-2023-ask}, \citet{deng-etal-2023-prompting} and \citet{Fu2023ImprovingLM} prompt LLMs to select strategies, relying on their parametric knowledge without additional training. In contrast, \citet{deng2024plug} and \citet{zhang2024strength} developed strategy planners based on small LMs via supervised fine-tuning and reinforcement learning. These methods enable dialogue agents to exhibit more proactive, goal-directed behavior and achieve strong performance across dialogue scenarios.

Despite the success, existing methods have several limitations. First, most of them rely on a set of pre-defined strategies that are relatively small in size. Such limited coverage of strategies constrains the agents' adaptability to diverse real-world scenarios \citep{nguyen2024dynasaur}. Second, recent work has highlighted that LLMs exhibit preference bias when selecting strategies \citep{kang-etal-2024-large}, which hinders their ability to identify optimal strategies. Third, many approaches focus on training external planners, which requires specifically curated datasets and may hinder generalization to unseen situations, thereby falling short of providing diverse and unbiased strategy planning.
As illustrated in Figure~\ref{fig:figure_challenge}, these limitations underscore the need for an alternative that expands strategy coverage and mitigates bias, while not relying on costly training.

To tackle these, we introduce \book \textbf{\Principles}: a synthetic
strategy memory for proactive dialogue agents, derived through offline self-play simulations. Inspired by how humans learn from both successful and failed experiences \cite{edmondson2011strategies, grossmann2017wisdom}, we leverage these two types of experiences to derive fundamental principles that enable effective strategy planning. 
Specifically, when the agent's strategy leads to success \textit{(e.g., resolving the user's core issue)}, we derive \Principles{} by analyzing the success factors. In contrast, when the strategy results in failure \textit{(e.g., worsening the user’s distress)}, we initiate a revision process where we backtrack to the starting point of the failure. From there, we iteratively revise the strategy and re-simulate the interaction until success. Then, we derive \Principles{} that capture what determines both successes and failures. 
This process allows us to accumulate strategies from both positive and negative experiences, structured as follows: \textit{\textbf{when} [situation], \textbf{you should} [successful strategy], \textbf{rather than} [failed strategies]\footnote{The \textit{\textbf{rather than}} clause is included only when \Principles{} is extracted from a revision process.}, \textbf{because} [reason]}. 

\Principles{} has advantages in three key aspects:  
\textbf{(i) Coverage}: Although derived from a limited set of simulations, our strategy space effectively covers diverse dialogue scenarios, addressing the limitations of pre-defined strategy sets in prior work. 
\textbf{(ii) Bias}: \Principles{} semantically captures contrasts between effective and ineffective strategies \textit{(i.e., “you should … rather than …”)}, which explicitly helps avoid dialogue agents' harmful bias toward improper strategies.
\textbf{(iii) Training}: By uncovering the hidden parametric knowledge of LLMs in a non-parametric form, \Principles{} enhances dialogue agents without additional training or reliance on vast amounts of human conversation data.

After demonstrating the effectiveness of \Principles{} over common baselines in two commonly used datasets of proactive dialogue (\eg, ESConv for emotional support and P4G for persuasion), we extend the evaluation to more challenging environments where a broader range of strategies is required. Specifically, we use ExTES \cite{zheng2023building} and construct an extended version of P4G.
Despite the increased complexity, our method achieves strong performance, demonstrating its effectiveness in more realistic dialogue contexts.

\begin{figure*}[t]
    \centering
    \includegraphics[width=1\linewidth]{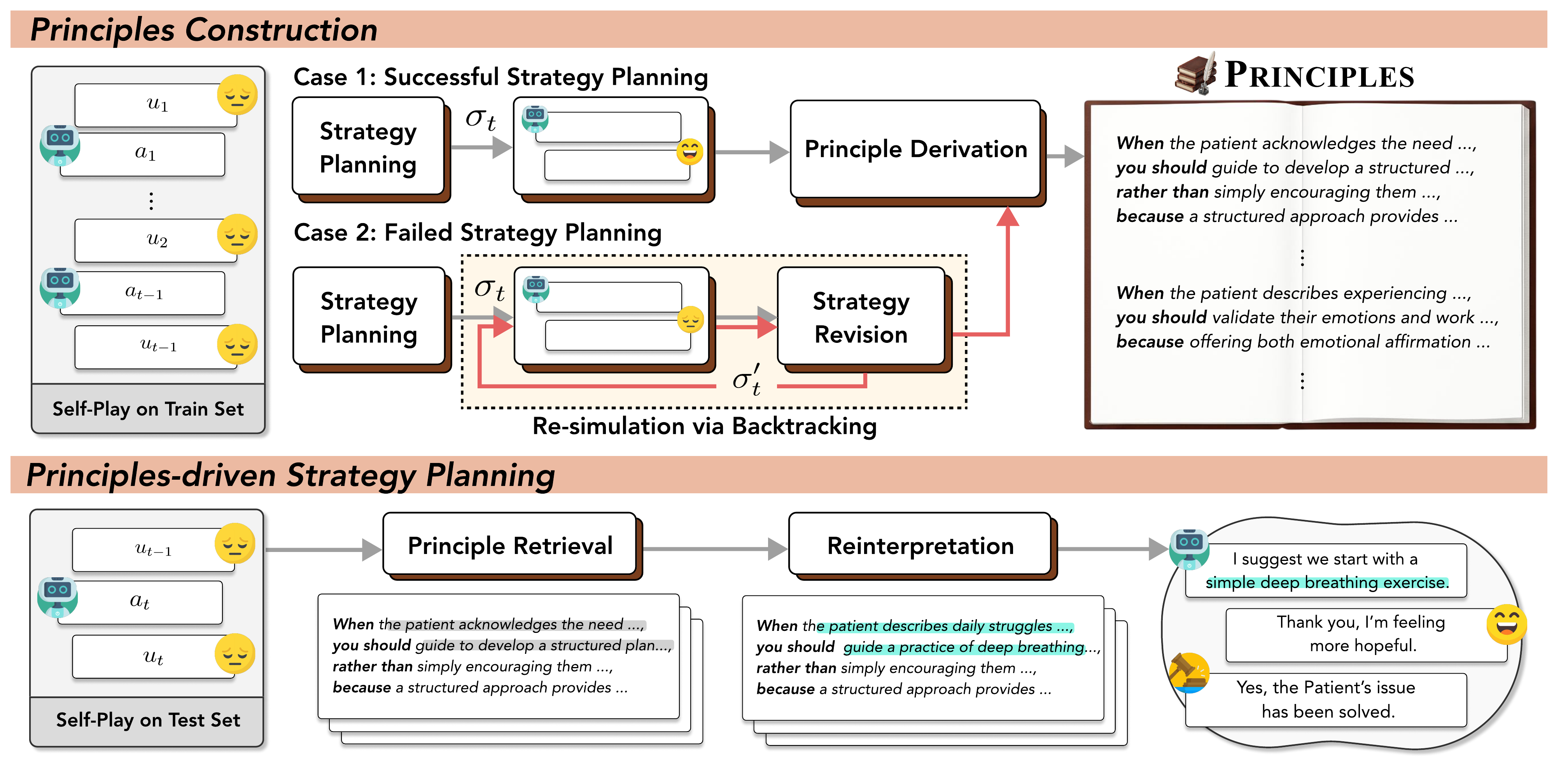}
    \caption{The overview of constructing \book \textbf{\Principles{}} and applying them to strategy planning. Top: principles construction via offline self-play simulations; Bottom: principles-driven strategy planning during inference.}
    \label{fig:main_figure}
\end{figure*}

\section{Preliminary: Self-Play Simulation}
\label{section:self-play}
Following \citet{deng2024plug}, we adopt self-play simulation to model strategic decision-making in proactive dialogue. In this setup, an agent engages in multi-turn conversations with a user simulator, adaptively selecting strategies at each turn and responding accordingly to accomplish a defined goal.

At each turn \( t \), the agent observes the current state \( s_t = \{ a_1, u_1, a_2, \ldots, a_{t-1}, u_{t-1} \}\), which consists of the dialogue history up to that point, \textit{i.e.,} all utterances from turns \( 1 \) to \( t{-}1 \).
Given \( s_t \), the agent selects a high-level strategy \( \sigma_t \in \Sigma \) either by prompting an LLM or using a tunable planner:
\begin{equation}
\sigma_t = \texttt{LLM}_{\theta}(\rho_{\sigma}; s_t) \quad \text{or} \quad \sigma_t = \pi_{\phi}(s_t)
\end{equation}
where \(\texttt{LLM}_{\theta}\) denotes a frozen LLM prompted with strategy selection instructions \(\rho_{\sigma}\), while \(\pi_\phi\) denotes a planner learned from data (\textit{e.g.,} via supervised fine-tuning or reinforcement learning).
Once a strategy \(\sigma_t\) is selected, the agent generates a response conditioned on the \( \sigma_t \) and \( s_t \):
\begin{equation}
a_t = \texttt{LLM}_{\theta}(\rho_{a}; \sigma_t, s_t)
\end{equation}
Subsequently, the user simulator generates a response based on \( s_t \) and \( a_t \):
\begin{equation}
u_t = \texttt{LLM}_{\theta}(\rho_{u}; s_t, a_t)
\end{equation}
where \(\rho_{a}\) and \(\rho_{u}\) are fixed role-specific prompts used to guide the agent and user simulator, respectively, during the self-play simulation.
Finally, verbal feedback is generated by a critic model \(\texttt{LLM}_{\theta}\). This feedback is then mapped to a scalar reward via a fixed mapping function \(f(\cdot)\), \textit{e.g.,} the verbal output “The patient’s issue has been solved.” would be mapped to a score of 1.0.
To reduce variance from stochastic decoding, the final reward \(r_t\) is obtained by averaging over \(l\) sampled outputs:
\begin{equation}
r_t = \frac{1}{l} \sum_{i=1}^{l} f\left(\texttt{LLM}_{\theta}^{(i)}(\rho_{c}; s_t, a_t, u_t)\right)
\end{equation}
where \(\rho_c\) is a prompt that elicits verbal feedback (see Appendix~\ref{appendix:evaluation} for details).

\section{\Principles}

Inspired by \citet{louie2024roleplay}, which elicits qualitative feedback from a domain expert, we propose \book \textbf{\Principles{}}: a synthetic strategy memory derived from offline self-play simulations. We explain how \Principles{} is constructed and how it can be applied in real-time conversation (Figure~\ref{fig:main_figure}).

\subsection{Principles Construction}
\paragraph{Success and Failure Detection.}
To enable dialogue agents to learn from both success and failure, we start with a offline self-play simulations where we collect the agent's success and failure. 
At each turn \( t \), the agent and the user simulator generate their responses, and a critic model assigns a scalar reward \( r_t \).  
We determine the \texttt{status} as either success or failure by evaluating whether the reward is higher than the previous turn:
\begin{equation}
\texttt{status}(s_t, a_t, u_t) = 
\begin{cases}
\texttt{1} & \text{if } r_t > r_{t-1} \\
\texttt{0} & \text{otherwise}
\end{cases}
\end{equation}
If the turn is successful (\texttt{status} == \texttt{1}), we immediately derive a principle \( p_t \) based on the agent's strategy and the dialogue context:
\begin{equation}
p_t = \texttt{LLM}_{\theta}(\rho_{\pi}; s_{t}, \mathcal{T}_t)
\end{equation}
where \( \rho_{\pi} \) is a prompt designed to extract a principle from a successful case, and \( \mathcal{T}_t = (\sigma_t, a_t, u_t) \) denotes the successful interaction consisting of the agent’s strategy \(\sigma_{t}\), utterance \(a_t\), and user response \(u_t\) at turn \(t \). We then add the resulting principle \(p_t\) to the set \(\mathcal{P}\) of accumulated \Principles:
\begin{equation}
\mathcal{P} \leftarrow \mathcal{P} \cup \{ p_t \}
\end{equation}
Finally, the state transitions to the next turn, where \( s_{t+1} = \{a_1, u_1, \ldots, a_t, u_t\} \). 
Unlike the successful case where a principle is immediately extracted, if the turn fails (\texttt{status} == \texttt{0}), we invoke a strategy revision process described in the following section.

\paragraph{Strategy Revision.}
Upon detecting a failure, the simulation invokes a revision step to refine the previously failed strategic decision.  
It then generates a revised strategy \(\sigma_t^{\prime}\) to re-simulate from the failure point, leveraging prior failed attempts at turn \(t\). Formally, the revised strategy is generated as:
\begin{equation}
\sigma_t^{\prime} = \texttt{LLM}_{\theta}(\rho_{r}; s_t, \mathcal{F}_t)
\end{equation}
where \(\rho_{r}\) is the revision prompt and \(\mathcal{F}_t\) denotes the set of previously failed trials at turn \(t\), defined as \( \mathcal{F}_t = \{ (\sigma_t^{1}, a_t^{1}, u_t^{1}), \dots, (\sigma_t^{n}, a_t^{n}, u_t^{n}) \} \)
where \(n\) is the maximum number of failed attempts. This failure history guides the model to avoid previously ineffective strategies.

\paragraph{Re-simulation via Backtracking.}
After generating a revised strategy \(\sigma_t^{\prime}\), the simulation backtracks to the original state \(s_t\) preceding the failure and re-simulates turn \(t\) using \(\sigma_t^{\prime}\). The agent generates a revised response \(a_t^{\prime}\), and the user simulator produces a new reply \(u_t^{\prime}\) based on the updated context.
\begin{equation}
a_t^{\prime} = \texttt{LLM}_{\theta}(\rho_{a}; s_t, \sigma_t^{\prime}) 
\end{equation}
\begin{equation}
u_t^{\prime} = \texttt{LLM}_{\theta}(\rho_{u}; s_t, a_t^{\prime})
\end{equation}
The resulting interaction is then evaluated by the critic model to compute a revised scalar reward \(r_t'\). This process is repeated until either a successful outcome is found or a maximum number of attempts is reached.

\paragraph{Principle Derivation.}
If the corrected turn is re-evaluated as successful (\texttt{status} == \texttt{1}), indicating a transition from failure to success, we derive a principle \( \tilde{p_t} \) as a result of overcoming the failure:
\begin{equation}
\tilde{p_t} = \texttt{LLM}_{\theta}(\rho_{\psi}; s_{t}, \mathcal{T}_t^{*}, \mathcal{F}_{t})
\end{equation}
where \(\rho_{\psi}\) is a prompt designed to extract a principle from failure, and the successful revised interaction is denoted as \(\mathcal{T}_t^{*} = (\sigma_t^{*}, a_t^{*}, u_t^{*})\). The extracted principle is then added to the principle set \(\mathcal{P}\):
\begin{equation}
\mathcal{P} \leftarrow \mathcal{P} \cup \{ \tilde{p_t} \}
\end{equation}
Each principle is represented in a structured format to ensure interpretability and reusability:
\begin{list}{}{\leftmargin=2.1em}
\item
\ttfamily
\textbf{When} [situation], \\
\textbf{you should} [successful strategy], \\
\textbf{rather than} [failed strategy], \\
\textbf{because} [reason].
\end{list}
Then, the state transitions to the next state \( s_{t+1} = \{a_1, u_1, \ldots, a_t^{*}, u_t^{*}\} \), incorporating the successfully revised turn.
Consequently, this process uncovers hidden parametric knowledge---previously inaccessible due to model bias---by iteratively identifying failures and refining strategies, ultimately transforming it into a non-parametric form. 

\begin{table*}[t]
    \centering
    \resizebox{1\textwidth}{!}{
    \begin{tabular}{l c cc cc cc cc}
        \toprule
        \multirow[c]{2}{*}{\textbf{Method}} & \multirow[c]{2}{*}{\textbf{$|\mathcal{S}|$}}
        & \multicolumn{2}{c}{\textbf{ESConv}} 
        & \multicolumn{2}{c}{\textbf{ExTES}} 
        & \multicolumn{2}{c}{\textbf{P4G}} 
        & \multicolumn{2}{c}{\textbf{P4G\textsuperscript{+}}} \\
        \cmidrule(lr){3-4} \cmidrule(lr){5-6} \cmidrule(lr){7-8} \cmidrule(lr){9-10}
        & & SR$\uparrow$ & AT$\downarrow$ 
          & SR$\uparrow$ & AT$\downarrow$ 
          & SR$\uparrow$ & AT$\downarrow$ 
          & SR$\uparrow$ & AT$\downarrow$ \\
        \midrule
        Standard & & & & & & & & & \\
        \quad + GPT-3.5-Turbo & 0 & 0.4154 & 8.59 & 0.4923 & 8.15 & 0.8000 & 5.41 & 0.3667 & 7.78 \\
        \quad + GPT-4o & 0 & 0.5583 & 8.13 & 0.6667 & 7.30 & 0.9375 & \underline{4.07} & 0.4917 & 7.14 \\
        \midrule
        
        Proactive \cite{deng-etal-2023-prompting} & 8-16 & 0.2385 & 9.51 & 0.5615 & 8.24 & 0.9500 & 4.23 & 0.4333 & 7.35 \\
        \ + MI-Prompt \cite{chen-etal-2023-controllable} & 8-16 & 0.3769 & 8.93 & 0.6538 & 7.82 & 0.9083 & 4.18 & 0.3417 & 7.91 \\
        ProCoT \cite{deng-etal-2023-prompting} & 8-16 & 0.2231 & 9.51 & 0.6308 & 7.99 & \underline{0.9583} & 4.15 & 0.4500 & 7.24 \\
        \ + MI-Prompt \cite{chen-etal-2023-controllable} & 8-16 & 0.3538 & 9.13 & 0.6692 & 7.58 & 0.9250 & \textbf{3.66} & 0.4333 & 7.44 \\
        PPDPP \cite{deng2024plug} & 8-16 & 0.5077 & 8.16 & 0.6846 & 6.99 & \textbf{0.9667} & 4.41 & -- & -- \\
        \midrule
        AnE \cite{zhang-etal-2024-ask} & $\infty$  & \underline{0.5846} & \underline{7.38} & 0.6462 & \underline{6.93} & 0.9083 & 4.27 & \underline{0.5333} & \underline{6.78} \\
        ICL-AIF \cite{Fu2023ImprovingLM} & $\infty$  & 0.5615 & 7.87 & \underline{0.7154} & 7.37 & 0.8000 & 4.68 & 0.5083 & \textbf{6.70} \\

        \midrule
        \midrule
        \textbf{\book \Principles \ (Ours)} & $\sim$100 & \underline{0.7385} & \textbf{6.36} & \textbf{0.8615} & \textbf{5.87} & 0.9500 & 4.73 & \textbf{0.5917} & 7.15 \\
        \bottomrule
    \end{tabular}
    } 
    \caption{Comparison of performance across four proactive dialogue tasks via self-play simulations. Here, $|\mathcal{S}|$ denotes the size of the strategy set used in each setting.}
    \label{tab:table_1}
    \vspace{-0.3cm}
\end{table*}


\subsection{Principles-driven Strategy Planning}
\paragraph{Retrieval and Reinterpretation.}
To apply the extracted \Principles{} during inference, we first identify candidate principles that closely match the current context. Since the \textbf{\texttt{When}} clause captures the core situation, we retrieve relevant top-\(k\) principles by comparing the current state \(s_t\) and the \texttt{When} clause using L2 distance between embedding vectors.
Only the \textbf{\texttt{When}} component of each principle is used to compute similarity, allowing the agent to identify contextually analogous dialogue situations across diverse scenarios. We denote the set of top-\(k\) retrieved principles as \( \Sigma_t = \{\sigma_1, \dots, \sigma_k\} \subset \mathcal{P} \). Since even within the same domain, retrieved principles may not directly align with the dialogue context, we perform a reinterpretation step. Formally, the reinterpreted principles \(\tilde{\Sigma}_t\) are generated as:
\begin{equation}
\tilde{\Sigma}_t = \texttt{LLM}_{\theta}(\rho_{\nu}; s_t, \Sigma_t)
\end{equation}
where \(\rho_{\nu}\) is a reinterpretation prompt designed to adapt retrieved principles \(\Sigma_t\) to the current context. This aligns each principle with the context.

\subsection{Implementation Details}
We implement our approach based on the construction and planning methods described earlier. Below, we detail how \Principles{} is constructed and applied in practice.

\paragraph{Principle Construction.}
To construct our \Principles{}, we adopt the self-play simulation scheme introduced in Section~\ref{section:self-play}, following prior work \citep{deng2024plug}. For each dataset, we run 50 simulations, initialized with the first turn of a dialogue from the training set, which is used solely to expose the model to diverse situations for self-play rather than to reproduce full conversations.
The agent and the user simulator interact for up to 10 turns, guided by role-specific prompts, while a critic model assigns scalar rewards after each turn. A principle is derived when the reward improves over the previous turn; otherwise, a revision process is triggered, with up to 3 attempts to avoid getting stuck in failure loop. On average, about 100 principles are derived from 50 simulations per domain. See Appendix~\ref{appendix:algorithm} and ~\ref{appendix:prompt_for_ours} for the detailed algorithm and prompt templates.

\paragraph{Principles-driven Strategy Planning.}
During inference, we retrieve the top-$k$ most relevant principles (default: 3), based on the L2 distance between embedding vectors. Specifically, we embed the current state and the \textbf{\texttt{When}} clause of each principle using OpenAI’s embedding model (\textit{i.e.,} \texttt{text-embedding-ada-002}). Retrieval is implemented using the \texttt{FAISS} library \citep{douze2025faisslibrary}. To align the retrieved principles with the current state, we employ a reinterpretation process, with prompting details provided in Appendix~\ref{appendix:prompt_for_ours}.

\paragraph{Large Language Models.}
We find prior work's effectiveness is often overestimated due to limited evaluation settings. They focus on an earlier-generation LLM (\textit{e.g.,} \texttt{gpt-3.5-turbo}) as both user simulator and critic. 
To ensure more robust evaluation, we adopt a newer model (\textit{i.e.,} \texttt{gpt-4o}), 
which applies stricter criteria by requiring resolution of the user’s core issue rather than surface-level relief. 
Human evaluation further confirms \texttt{gpt-4o}’s closer alignment with human judgments.
This demonstrates that our evaluation setting is more reliable 
(see Appendix~\ref{appendix:comparison_critic} for details).
\section{Evaluation}

\subsection{Evaluation Setups}
\paragraph{Evaluation Metrics.} We use two core metrics commonly adopted in strategy planning: success rate and average turns. In addition, to more comprehensively evaluate, we include three automatic metrics: macro F1, weighted F1, and entropy.

\begin{itemize}
    \item \textbf{Success Rate} (SR): Success is determined by whether the reward assigned after each turn exceeds a threshold.
    
    \item \textbf{Average Turns} (AT): The average number of turns across all episodes. This reflects how efficiently the agent can achieve the task goal.

    \item \textbf{Macro F1 Score \( (F_{m})\)}: Evaluates the model’s alignment with human-annotated strategy labels, by averaging per-class F1 scores.

    \item \textbf{Weighted F1 Score \( (F_{w}) \) }: Computes the average of per-class F1 scores weighted by class frequency, mitigating imbalance.

    \item \textbf{Entropy \( (H) \)}: This measures the diversity of predicted strategies, where higher entropy indicates lower bias toward specific strategies.
    {\small
    \begin{equation}
    H = - \sum_{c=1}^{C} p_c \log p_c
    \end{equation}
    }
\end{itemize}

\paragraph{Baselines.}
We first evaluate a standard agent that operates without explicit strategy guidance. 
We then compare against two types of baselines. 
\textbf{(i)} \underline{Using pre-defined strategies}, such as Proactive \cite{deng-etal-2023-prompting} and ProCoT \cite{deng-etal-2023-prompting}, which prompt the model to select from a limited set of strategies. 
In contrast, PPDPP \citep{deng2024plug} employs a lightweight external planner trained via supervised fine-tuning and reinforcement learning. 
\textbf{(ii)} \underline{Using open-ended strategies}, such as Ask-an-Expert \citep{zhang-etal-2024-ask} and ICL-AIF \citep{Fu2023ImprovingLM}, which dynamically generate strategies using LLMs as expert knowledge sources or via AI feedback. 
This allows us to assess our approach against baselines with different levels of coverage and flexibility 
(see Appendix~\ref{appendix:baselines}).

\paragraph{Datasets.}
We first evaluate on ESConv and P4G, then extend to more challenging settings with ExTES and P4G\textsuperscript{+}. 
P4G\textsuperscript{+} extends the original task by incorporating (i) diverse personas, (ii) multiple organizations, and (iii) donation barriers (\textit{e.g.,} financial constraints), yielding more realistic persuasion scenarios 
(see Appendix~\ref{appendix:p4gplus} for details).

\subsection{Results}
In this section, we investigate our method’s performance in addressing three key challenges in strategy planning: coverage, bias, and training.

\paragraph{RQ1. Can \Principles{} effectively expand strategy coverage?}

As shown in Table~\ref{tab:table_1}, we reveal distinct performance trends across tasks depending on the size of a strategy set. Notably, with the improved capabilities of \texttt{gpt-4o} over the previous model, even the standard agent without any explicit strategy use achieves competitive SR in most tasks.
Interestingly, methods with open-ended settings (\textit{i.e.,} AnE, ICL-AIF) achieve higher SR on average than baselines relying on a pre-defined strategy set. Furthermore, even the standard baseline occasionally outperforms them.
This suggests that constraining the model to a fixed set of strategies may hinder its capacity for proactive dialogue. A notable exception is P4G, where the task is focused on a single goal (\textit{i.e.,} donating to Save the Children), making it possible for a concise strategy set to sufficiently cover the task.
In contrast, we construct \Principles \ that efficiently expand the strategy coverage using only 50 offline self-play simulations on the training set. With these resources, our strategy planning consistently achieves superior performance across a range of tasks.

\begin{table}[t]
\centering
\resizebox{1\columnwidth}{!}{
\begin{tabular}{@{}lcccccc@{}}
\toprule
\multirow[c]{2}{*}{\textbf{Method}}
& \multicolumn{3}{c}{\textbf{ESConv}} 
& \multicolumn{3}{c}{\textbf{P4G}} \\
\cmidrule(lr){2-4} \cmidrule(lr){5-7}
& \textbf{\( F_m \)} & \textbf{\( F_w \)} & \textbf{\( H \)} 
& \textbf{\( F_m \)} & \textbf{\( F_w \)} & \textbf{\( H \)} \\
\midrule
Proactive & 6.91 & 9.25 & 0.87 & 13.95 & 21.58 & \underline{2.60} \\
ProCoT & 6.94 & 9.85 & 0.90 & 13.73 & 20.50 & \underline{2.60} \\
PPDPP & 6.15 & 11.33 & 0.07 & 7.91 & 10.66 & 1.03 \\
AnE & \underline{7.78} & \underline{13.76} & \underline{1.07} & \underline{14.65} & \underline{23.27} & \underline{2.60} \\
ICL-AIF & 4.61 & 10.69 & 0.11 & 10.79 & 16.74 & 2.46 \\
\midrule

\textbf{Ours} & \textbf{10.52} & \textbf{17.67} & \textbf{1.21} & \textbf{14.96} & \textbf{24.30} & \textbf{2.67} \\
\bottomrule
\end{tabular}
}
\caption{Evaluation of strategy prediction performance and distributional diversity on ESConv and P4G.}
\label{tab:table_2}
\end{table}

\begin{table}[t]
    \centering
    \scalebox{1}{ 
    \resizebox{1\columnwidth}{!}{
    \begin{tabular}{@{}l cc cc@{}}
        \toprule
        \multirow[c]{2}{*}{\textbf{Method}}
        & \multicolumn{2}{c}{\textbf{ExTES}} 
        & \multicolumn{2}{c}{\textbf{P4G\textsuperscript{+}}} \\
        \cmidrule(lr){2-3} \cmidrule(lr){4-5}
        & SR$\uparrow$ & AT$\downarrow$ 
        & SR$\uparrow$ & AT$\downarrow$ \\
        \midrule
        \textbf{\book \Principles \ (Ours)} & \textbf{0.8615} & \underline{5.87} & \textbf{0.5917} & 7.15 \\
        \midrule
        w/o Structured & \underline{0.8385} & \textbf{5.33} & \underline{0.5667} & \textbf{6.39}  \\
        w/o Retrieval & 0.7846 & 5.91 & 0.5833 & 6.62  \\
        w/o Reinterpretation & \underline{0.8385} & 5.84 & \underline{0.5667} & \underline{6.49}  \\
        \bottomrule
    \end{tabular}
    }
    }
    \caption{Performance of our ablations.}
    \label{tab:table_3}
\end{table}

\paragraph{RQ2. Can strategy planning based on \Principles \ help mitigate strategy bias?}

To evaluate whether \Principles{} reduce preference bias in strategy planning, we assess strategy prediction performance on ESConv and P4G, where human-annotated strategy labels are available.
To ensure fair comparison with other methods, we allow free-form strategy generation in open-ended baselines and subsequently map each strategy to the closest pre-defined label using \texttt{gpt-4o}.

In Table~\ref{tab:table_2}, our method shows the highest macro F1, weighted F1, and entropy, indicating that it not only selects contextually appropriate strategies but also avoids overusing preferred strategies. In contrast, baselines often exhibit strong bias, consistently relying on a few dominant strategies. For example, PPDPP heavily overfits one or two strategies during reinforcement learning, resulting in a highly skewed distribution (Appendix~\ref{appendix:bias}).
These findings are supported by our ablation studies in Table~\ref{tab:table_3}. When \Principles{} are extracted in an unstructured format, or when similarity-based retrieval is replaced by selecting via LLM, performance consistently degrades. These results highlight that the structured format (\textit{i.e., “When… you should… rather than…”}) and retrieval process guide the model to select contextually relevant strategies instead of sticking to biased preferences, improving downstream performance.

\begin{figure}[t]
    \centering
    \includegraphics[width=0.85\linewidth]{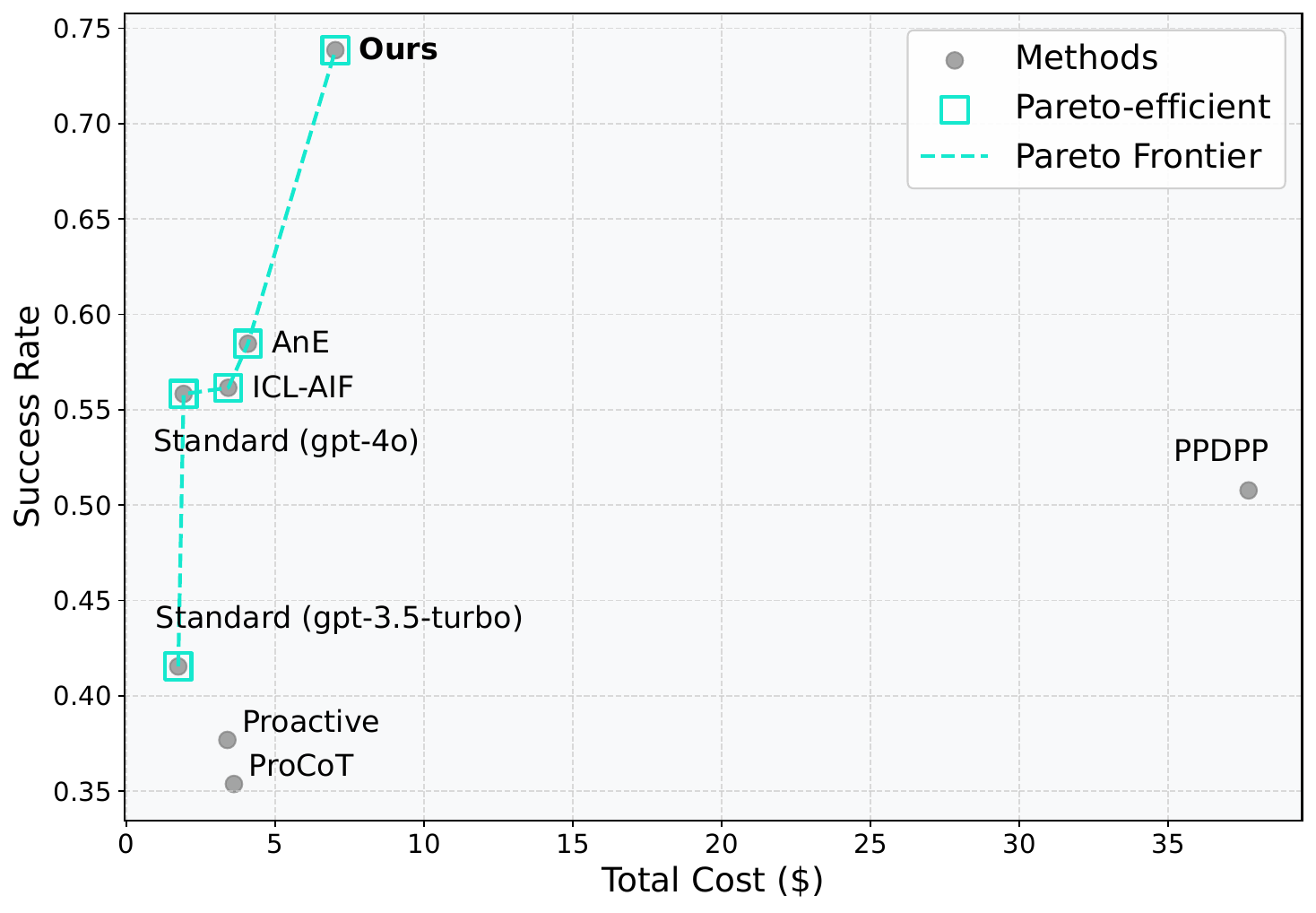}
    \caption{Cost-performance comparisons.}
    \label{fig:figure_pareto}
\end{figure}

\begin{figure*}[t]
    \centering
    \includegraphics[width=1\linewidth]{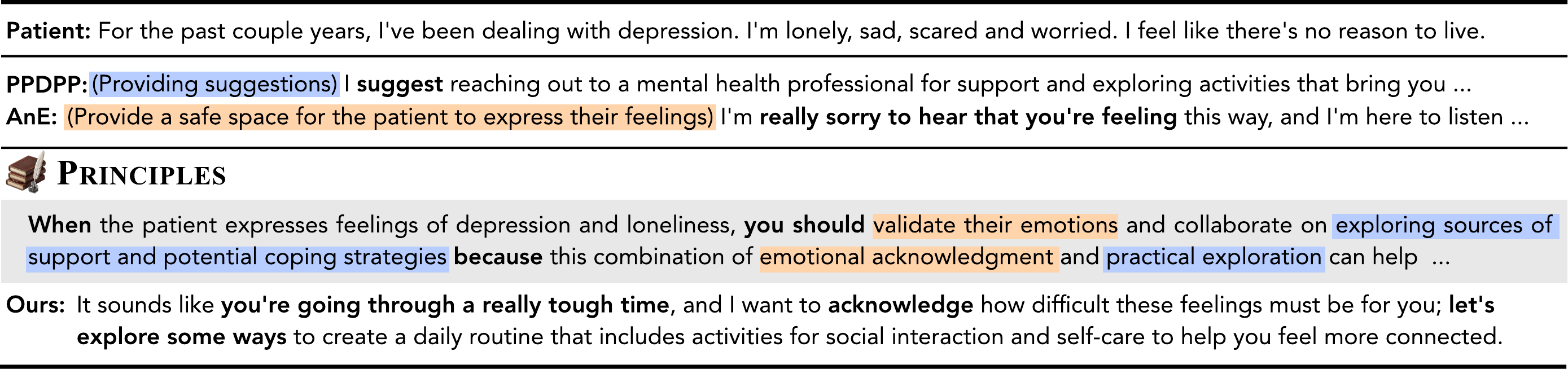}
    \caption{Qualitative example comparing AnE, PPDPP, and our approach based on \Principles.}
    \label{fig:figure_case}
\end{figure*}

\paragraph{RQ3. Can \Principles \ improve agents without training?}

Unlike prior methods that rely on fine-tuning \citep{deng2024plug, he2025simulating, he2025simulation}, we effectively address coverage and bias issues without any additional training.
Figure~\ref{fig:figure_pareto} compares the training and inference costs of ours against baselines, including PPDPP, which involves SFT in human-labeled data and RL via 1,000 self-play simulations. As a result, PPDPP incurs a significantly higher training cost (\textit{i.e.,} approximately 11.5$\times$ more than our method).
In contrast, our approach requires no labeled data and performs only 50 offline simulations on the training set, maintaining a cost comparable to prompt-based baselines while consistently outperforming all of them.
These findings highlight that the model already possesses extensive knowledge for strategy planning, and that unlocking and leveraging this \textbf{\textit{hidden parametric knowledge}} is important to drive strong performance improvements rather than updating model parameters.

\section{Human Evaluation}

\begin{figure}[h]
    \centering
    \includegraphics[width=1\linewidth]{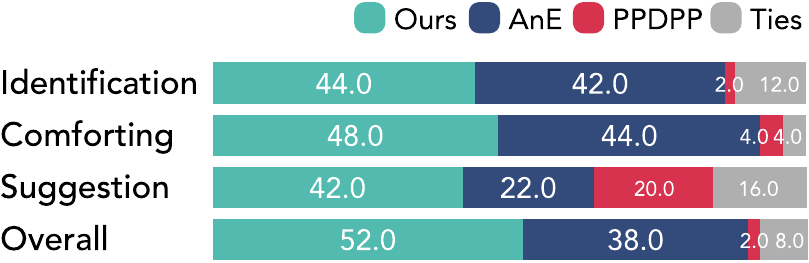}
    \caption{Human evaluation of response quality.}
    \label{fig:figure_human_eval}
\end{figure}
\paragraph{Evaluating Response Quality.}
Given the nature of self-play simulations, we evaluate responses generated from the same context (\textit{i.e.,} first user utterance) to ensure a fair comparison across baselines, instead of comparing full dialogues that may vary in length and flow.
We recruit three annotators to evaluate the quality of generated responses on 50 randomly sampled dialogue contexts from the ExTES, comparing outputs from three methods (\textit{i.e.,} AnE, using open-ended strategies; PPDPP, using pre-defined strategies; and Ours). To reduce position bias, all responses are presented to each annotator in shuffled order. In this setting, we conduct the evaluation based on four evaluation criteria \cite{liu2021towards}: 
(1) \textbf{Identification}: Which response explored the patient's situation more in depth and was more helpful in identifying their problems?
(2) \textbf{Comforting}: Which response was more skillful in comforting the patient?
(3) \textbf{Suggestion}: Which response gave more helpful suggestions for the patient’s problems?
(4) \textbf{Overall}: Generally, which response's emotional support do you prefer?

As shown in Figure~\ref{fig:figure_human_eval}, our method consistently outperforms both baselines across all four criteria. This result can be further interpreted in conjunction with Figure~\ref{fig:figure_case}. We observe that AnE tends to overly reflect the patient's feelings without addressing the core issue, while PPDPP is biased toward providing suggestions, leading to interactions that repeat similar utterances without adjusting to the conversational flow. In contrast, our method achieves a more effective balance between emotional empathy and problem-solving by retrieving contextually appropriate strategies. This results in responses that align more closely with human preferences.

\begin{figure}[h]
    \centering
    \includegraphics[width=0.9\linewidth]{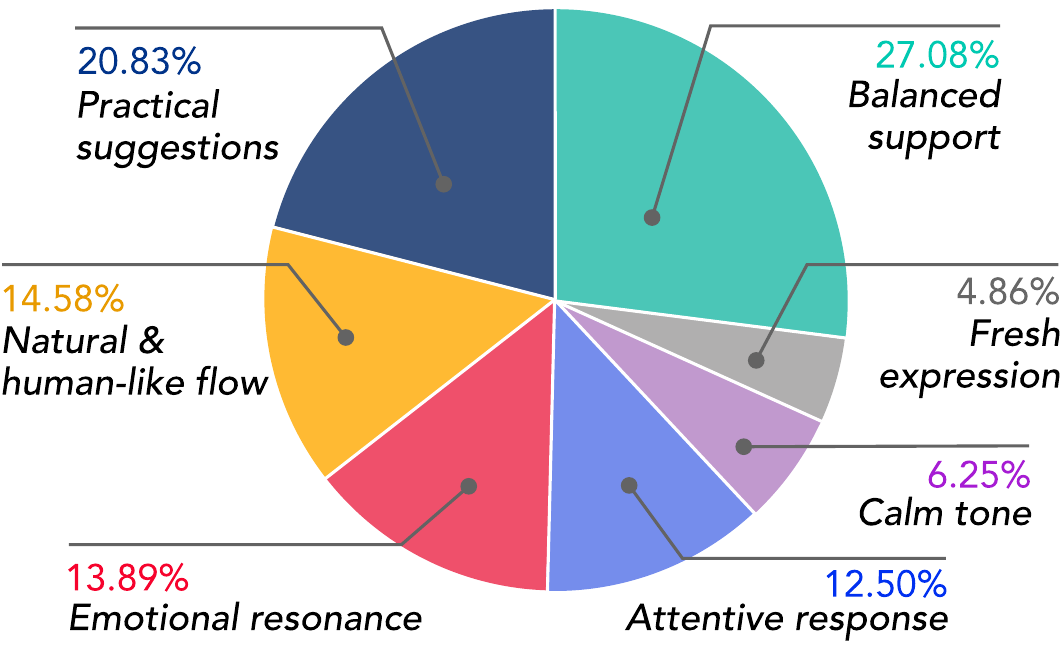}
    \caption{Analysis of human preference for our method.}
    \label{fig:figure_human_eval_2}
\end{figure}

\paragraph{Analysis of Human Preference.}

To gain deeper insight into human preferences, we asked participants to select reasons for their choice into several thematic categories. As shown in Figure~\ref{fig:figure_human_eval_2}, our approach tended to combine logical coherence and emotional empathy (\textit{i.e.,} Balanced Support).

\section{In-Depth Analysis}

\begin{table}[b]
\centering
\resizebox{0.49\textwidth}{!}{
\begin{tabular}{lcccccc}
\toprule
\textbf{Method} & $|\mathcal{S}|$ & SR$\uparrow$ & AT$\downarrow$ & Cost$_{\text{train}}$ & Cost$_{\text{infer}}$ & Time$_\text{infer}$ \\
\midrule
DPDP        & 8--16   & \textbf{0.7923} & 7.52 & \$59.44 & \$16.29 & 81.4s \\
\textbf{Ours} & $\sim$100 & 0.7385 & \textbf{6.36} & \textbf{\$3.29}  & \textbf{\$5.30}  & \textbf{30.5s} \\
\bottomrule
\end{tabular}
}
\caption{Comparison between DPDP and our method on the ESConv dataset.}
\label{tab:dpdp_comparison}
\end{table}

\paragraph{Comparison with Strong Baseline.}
To further validate our approach, we compare it with DPDP~\citep{he-etal-2024-planning}, a strong baseline using Monte Carlo Tree Search (MCTS)-based strategy selection at each turn. As shown in Table~\ref{tab:dpdp_comparison}, our method achieves competitive SR while attaining better performance in AT. In terms of efficiency, DPDP requires $18.07\times$ higher training cost than our principles construction, along with $3.07\times$ higher inference cost and $2.67\times$ longer inference time. While DPDP achieves slightly higher SR, its substantial computational overhead limits practicality, whereas our method provides a more efficient and scalable alternative for real-world applications.

\begin{figure}[t]
    \centering
    \includegraphics[width=0.95\linewidth]{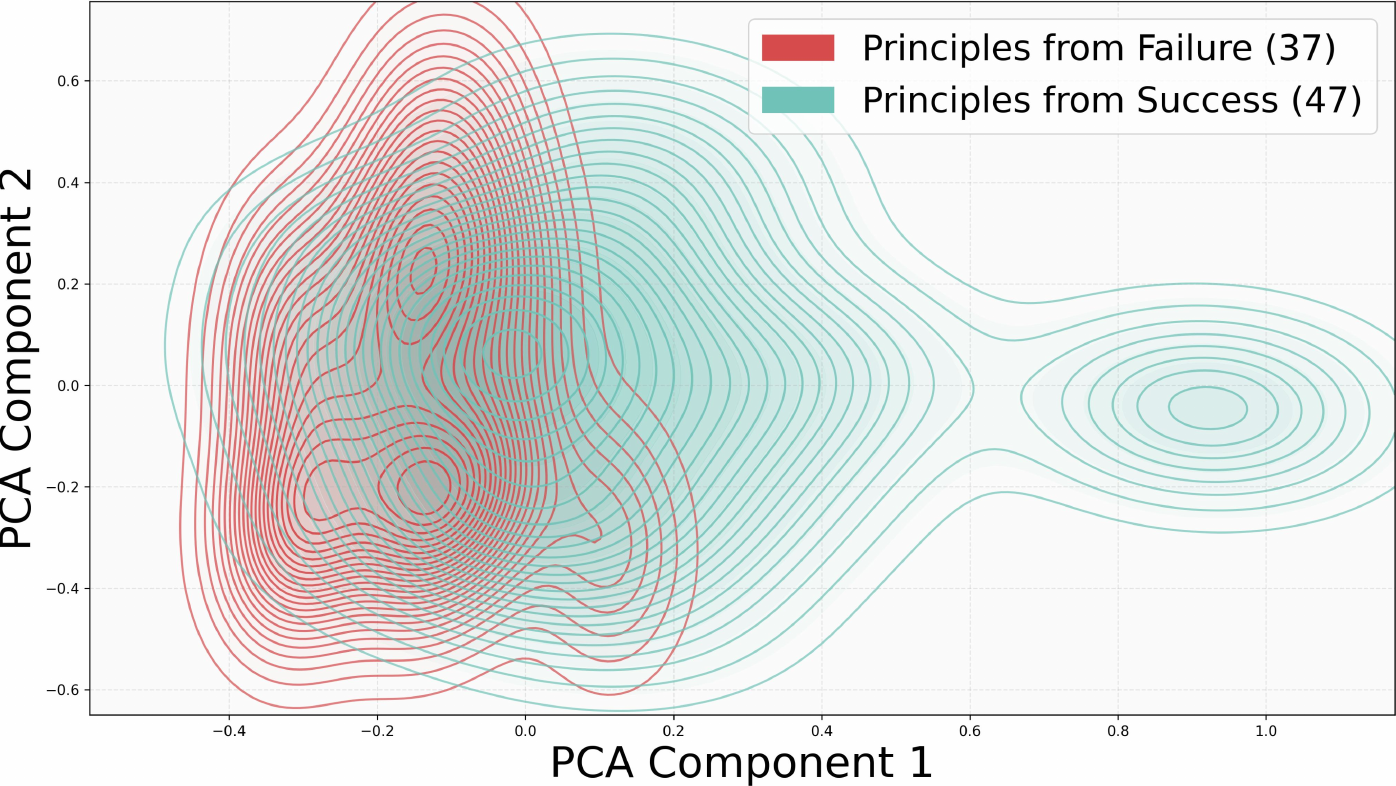}
    \caption{PCA projection of \Principles{} derived from successful and failed interactions. The distributions indicate that both contribute complementary strategic coverage.}
    \label{fig:figure_distribution}
\end{figure}

\begin{figure}[t]
    \centering
    \includegraphics[width=0.95\linewidth]{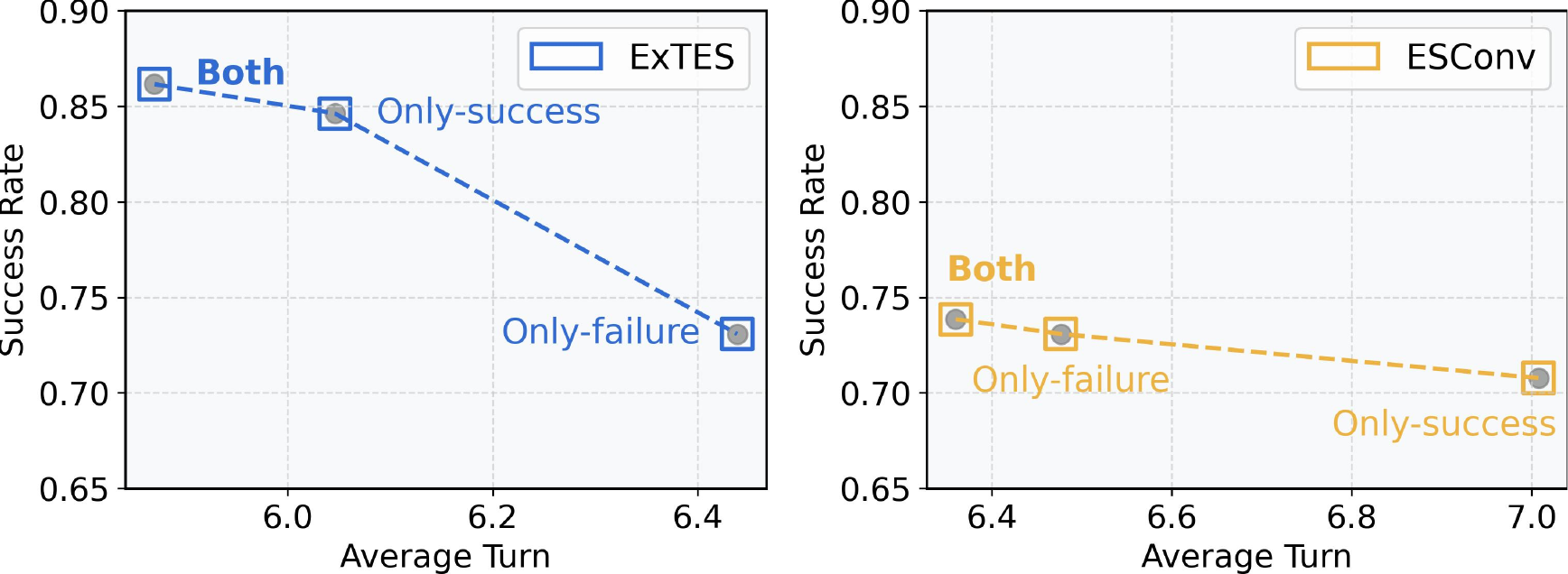}
    \caption{Comparison of performance using \Principles \ derived from success only, failure only, or both.}
    \label{fig:figure_both}
\end{figure}

\paragraph{Learning from Success, Failure, or Both.}
Figure~\ref{fig:figure_distribution} illustrates the effect of \Principles{} extracted from successful and failed interactions---an essential component of our method. We project the embedding vectors of these \Principles{} into a 2D space using Principal Component Analysis (PCA). While some overlap exists, each region clearly possesses its own distinct area of focus.
Additionally, in Figure~\ref{fig:figure_both}, we evaluate the practical impact of these two types of resources. On ExTES, \Principles{} derived from successful interaction lead to better outcomes than those extracted from failed ones, while the opposite is observed on ESConv. This indicates that neither source demonstrates consistent superiority. Instead, the integration of both types of \Principles{} yields the best performance, demonstrating the broader strategy coverage.

\paragraph{Diversity of Linguistic Organization Patterns.}
We further analyze the diversity of linguistic organization patterns used in principle construction. Interestingly, as shown in Table~\ref{tab:linguistic_formats}, we observe that removing even a single component from our original pattern results in a substantial performance drop (\textit{i.e.,} w/o \texttt{rather than} and w/o \texttt{because}). In contrast, an alternative format that retains all four components achieves slightly better SR (\textit{i.e.,} \texttt{If/then/instead of/in order to}), despite its surface-level differences. These findings suggest that the effectiveness of \Principles{} derives not from their superficial linguistic form but from the presence of four core informational elements: \textbf{the situation, the successful strategy, the failed strategy, and the rationale}. This insight provides a valuable design guideline for future research.

\begin{table}[t]
\centering
\resizebox{\columnwidth}{!}{
\begin{tabular}{lcc}
\toprule
\multirow{2}{*}{\textbf{Linguistic Format}} 
  & \multicolumn{2}{c}{\textbf{ESConv}} \\
\cmidrule(lr){2-3}
  & SR$\uparrow$ & AT$\downarrow$ \\
\midrule
\textbf{\book \Principles{}}   & \underline{0.7385} & \textbf{6.36} \\
\midrule
w/o \texttt{rather than} & 0.6231 & 7.20 \\
w/o \texttt{because}     & 0.6400 & 7.21 \\
\texttt{If/then/instead of/in order to} & \textbf{0.7538} & \underline{6.52} \\
\bottomrule
\end{tabular}
}
\caption{Comparison of performance across different linguistic formats on ESConv.}
\label{tab:linguistic_formats}
\end{table}

\paragraph{Online Construction Setting.}
While our framework primarily relies on offline construction, we also investigate an online construction setting (\textit{i.e.,} at inference) on ESConv dataset. In this setting, the agent derives \Principles{} only from successful interactions, since test-time interaction does not allow revisiting failed trajectories (\textit{i.e.,} no backtracking). As shown in Table~\ref{tab:online_offline}, online construction yields a moderate performance drop compared to the offline setting, yet still produces highly competitive results. This demonstrates that even without pre-constructed principles, our framework can rapidly adapt and expand its strategy coverage during deployment. Such adaptability is particularly important for real-world scenarios, where agents must continuously learn from limited successes without relying on offline self-play simulations.

\begin{table}[h]
\centering
\resizebox{\columnwidth}{!}{
\begin{tabular}{lcc}
\toprule
\multirow{2}{*}{\textbf{Method}} 
  & \multicolumn{2}{c}{\textbf{ESConv}} \\
\cmidrule(lr){2-3}
  & SR$\uparrow$ & AT$\downarrow$ \\
\midrule
Offline Construction & \textbf{0.7385} & \textbf{6.36} \\
Online Construction (\textit{i.e.,} at inference)  & 0.6615 & 7.22 \\
\bottomrule
\end{tabular}
}
\caption{Comparison of performance on ESConv under offline and online principle construction settings.}
\label{tab:online_offline}
\end{table}

\paragraph{Impact of Source Models.}
We further analyze how different LLMs affect \Principles{} quality. As shown in Table~\ref{tab:principle_source_models}, \texttt{Claude}- and \texttt{Llama}-based \Principles{} achieve higher SR, which correlates with their length, but not with utterance length, suggesting that more detailed guidance is beneficial. Importantly, Token$_p$ and Token$_u$ exhibit different orderings, indicating that the key factor in performance stems from specificity rather than verbosity.

\paragraph{Effect of Simulation Budget.}

\begin{figure}[t]
    \centering
    \includegraphics[width=0.95\linewidth]{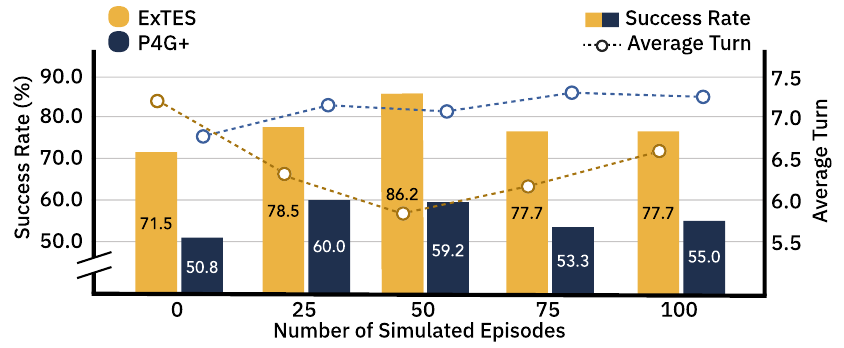}
    \caption{Correlation between a number of simulations and success rate.}
    \label{fig:figure_num_simulation}
\end{figure}

We investigate the optimal number of offline self-play simulations for effective principles construction. On both datasets, ExTES and P4G\textsuperscript{+}, we conduct 25, 50, 75, and 100 self-play simulations and measure SR and AT. As shown in Figure~\ref{fig:figure_num_simulation}, even 25 simulations yield substantial improvement, reaching its optimum at 50. However, the performance declines beyond 75, suggesting that principles that exceed the task's strategic requirements introduce additional noise that ultimately hinders overall model behavior.

\begin{figure}[t]
    \centering
    \includegraphics[width=0.95\linewidth]{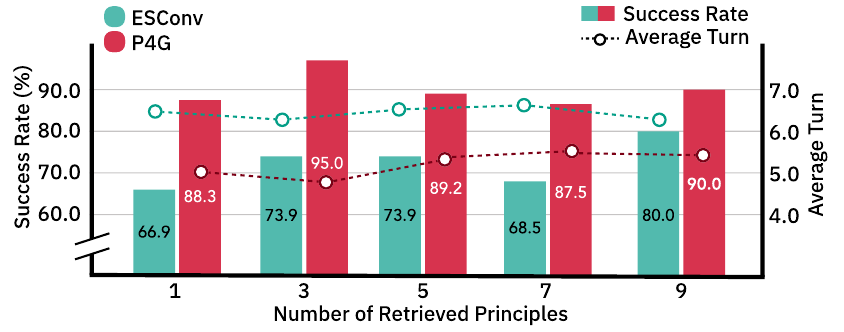}
    \caption{Correlation between a number of retrieved \Principles \ and success rate.}
    \label{fig:figure_topk}
\end{figure}

\paragraph{Impact of Top-k Strategies.}
We further investigate how the number of retrieved principles (\textit{k}) directly influences performance. 
As shown in Figure~\ref{fig:figure_topk}, performance varies with different values of \textit{k}, with the optimal number depending on the task (\textit{i.e.,} 9 for ESConv, 3 for P4G).
Notably, retrieving only a few top-ranked principles (\textit{e.g.,} top-3) still provides a highly competitive and cost-effective alternative in practical resource-constrained settings.

\begin{table}[t]
\centering
\resizebox{\columnwidth}{!}{
\begin{tabular}{lcccc}
\toprule
\multirow{2}{*}{\textbf{Source Model}} 
  & \multicolumn{4}{c}{\textbf{ESConv}} \\
\cmidrule(lr){2-5}
  & SR$\uparrow$ & AT$\downarrow$ & Tokens\(_p\) & Tokens\(_u\) \\
\midrule
\texttt{GPT-4o}            & 0.7385 & \underline{6.36} & 62.93 & 23.11 \\
\texttt{Claude-3.7-Sonnet} & \underline{0.7462} & 6.70 & 74.34 & 28.48 \\
\texttt{Llama-3.1-8B}      & \textbf{0.8615} & \textbf{5.44} & 83.34 & 22.93 \\
\bottomrule
\end{tabular}
}
\caption{Comparison of performance on ESConv with \Principles{} derived from different models. 
Tokens$_p$ denotes the average token length of the extracted \Principles{}, while Tokens$_u$ denotes the average token length of the agent’s utterances, guided by these \Principles{}.}
\label{tab:principle_source_models}
\end{table}

\section{Related Work}

Recent studies have increasingly highlighted the importance of strategy planning in proactive dialogue, such as emotional support \cite{liu2021towards} and persuasion \cite{wang2019persuasion}, where an agent should take the initiative to achieve a specific goal. To this end, many approaches rely on external planners to select the most suitable choice from a pre-defined strategy set \cite{deng-etal-2023-prompting}.
In contrast to methods that depend on a model’s parametric knowledge, \citet{deng2024plug} leverages human-annotated strategies via supervised fine-tuning (SFT), and further improves performance through reinforcement learning in simulated environments. Building upon prior work, \citet{zhang2024strength} and \citet{he2025simulating} incorporate user modeling to enable more tailored strategy selection.
Other approaches \citep{yu2023prompt, he-etal-2024-planning} formulate strategy planning as a search problem, using Monte Carlo Tree Search (MCTS) to identify optimal strategies. Another line of research focuses on open-ended strategy generation. For instance, \citet{Fu2023ImprovingLM} prompts LLMs to elicit improved strategies via iterative AI feedback, while \citet{zhang-etal-2024-ask} use LLMs as a source of expert knowledge. \citet{he2025simulation} attempts to eliminate the dependency on simulation environments by discovering latent policies from dialogue \cite{louie2024roleplay}.
Yet, they have several limitations, including limited coverage, preference bias, and costly training. This motivates the need of alternative approaches, which we introduce in our work.

\section{Conclusion}

In this paper, we introduce \book \textbf{\Principles{}}, a synthetic strategy memory for proactive dialogue agents, derived through offline self-play simulations.
Through extensive experiments across both standard and extended datasets, we show that our approach effectively expands strategy coverage and mitigates preference bias, leading to more balanced strategy planning.
Furthermore, our approach achieves promising performance without additional training by explicitly uncovering the hidden parametric knowledge into a structured, non-parametric form.
We expect our novel approaches to serve as a new foundation for future research efforts towards proactive dialogue agents.
\section*{Limitations}
First, our retrieval relies on embedding similarity over the \texttt{When} clause with the current state based on L2 distance, which may overlook subtle contextual nuances. Although our strategy planning approach based on \Principles{} includes a reinterpretation step to adapt retrieved principles to unseen scenarios, the selected principles may still fall short in highly specific or ambiguous dialogue situations, where fine-grained contextual understanding is required. One possible solution is to refine the scoring mechanism by combining embedding similarity with additional relevance signals, such as a dialogue stage, to improve retrieval accuracy beyond surface-level similarity.

Second, although our method supports turn-level planning via principle retrieval, it lacks explicit modeling of long-term goals. As a result, the agent may over-optimize for short-term reward, leading to suboptimal outcomes in tasks that demand long-term strategic planning, such as negotiations \citep{he2018decoupling}. Constructing principles from full dialogue trajectories may enhance long-term coherence, which we view as a promising direction to further develop our framework.
\section*{Ethical Considerations}
\paragraph{Human Annotation Process.} 
We recruit three undergraduate students with high English proficiency to evaluate the response quality. 
Evaluations are conducted on 50 randomly sampled dialogue contexts from the ExTES benchmark, covering outputs from three methods (AnE, PPDPP, and Ours). 
To guide consistency, we provide annotators with a custom-designed annotation form and detailed evaluation instructions (Figure~\ref{appendix:human_eval_ui}). 
To reduce position bias, all responses are shown in randomized order for each annotator.
\paragraph{LLM Usage and Ethical Risk.} We acknowledge that some of the datasets used (e.g., ESConv, ExTES) involve emotional support scenarios, and that the principles in our method are not curated by human experts. However, we do not intend to make any clinical or therapeutic claims regarding these principles. Our approach is intended purely as a research framework for exploring reusable dialogue strategies in simulation. Before any real-world deployment, especially in sensitive domains such as mental health, expert review and safety validation would be essential.
\section*{Acknowledgments}
This work was supported by Institute of Information \& Communications Technology Planning \& Evaluation (IITP) grant funded by the Korean government (MSIT) (No. RS-2020-II201361, Artificial Intelligence Graduate School Program (Yonsei University)), (No. RS-2024-00457882, National AI Research Lab Project), (2022-0-00077, RS-2022-II220077, AI Technology Development for Commonsense Extraction, Reasoning, and Inference from Heterogeneous Data). Jinyoung Yeo is the corresponding author.

\bibliography{reference}

\appendix

\begin{table*}[t]
    \centering
    \resizebox{0.8\textwidth}{!}{
    \begin{tabular}{l l cc cc cc cc}
        \toprule
        \multirow[c]{2}{*}{\textbf{Method}} & \multirow[c]{2}{*}{\textbf{Critic Model}}
        & \multicolumn{2}{c}{\textbf{ESConv}} 
        & \multicolumn{2}{c}{\textbf{ExTES}} 
        & \multicolumn{2}{c}{\textbf{P4G}} 
        & \multicolumn{2}{c}{\textbf{P4G\textsuperscript{+}}} \\
        \cmidrule(lr){3-4} \cmidrule(lr){5-6} \cmidrule(lr){7-8} \cmidrule(lr){9-10}
        & & SR$\uparrow$ & AT$\downarrow$ 
          & SR$\uparrow$ & AT$\downarrow$ 
          & SR$\uparrow$ & AT$\downarrow$ 
          & SR$\uparrow$ & AT$\downarrow$ \\
        \midrule
        \multirow{2}{*}{Standard} 
        & GPT-3.5-Turbo & 0.9154 & 4.52 & 0.9846 & 3.63 & 0.8583 & 4.73 & 0.5667 & 6.55 \\
        & GPT-4o         & 0.5583 & 8.13 & 0.6667 & 7.30 & 0.9375 & 4.07 & 0.4917 & 7.14 \\
        \bottomrule
    \end{tabular}
    }
    \caption{Comparison of performance between \texttt{gpt-3.5-turbo} and \texttt{gpt-4o} when used as critic models.}
    \label{appendix:table:critic}
\end{table*}

\begin{table*}[ht]
    \centering
        \resizebox{0.7\linewidth}{!}{
            \begin{tabular}{l|c}
            \hline
            \textbf{Human Rating} & \textbf{Count} \\
            \hline
            A. No, the Patient feels worse. & 1 \\
            B. No, the Patient feels the same. & 8 \\
            \textbf{C. No, but the Patient feels better. (GPT-4o's judgment)} & \textbf{54} \\
            D. Yes, the Patient's issue has been solved. (GPT-3.5's judgment) & 7 \\
            \hline
            \end{tabular}
        }
    \caption{Human ratings with model judgments.}
    \label{appendix:table:human_rating}
\end{table*}

\begin{figure*}[htbp]
    \centering
    \begin{subfigure}[b]{0.48\textwidth}
        \centering

        \includegraphics[width=1.05\textwidth]{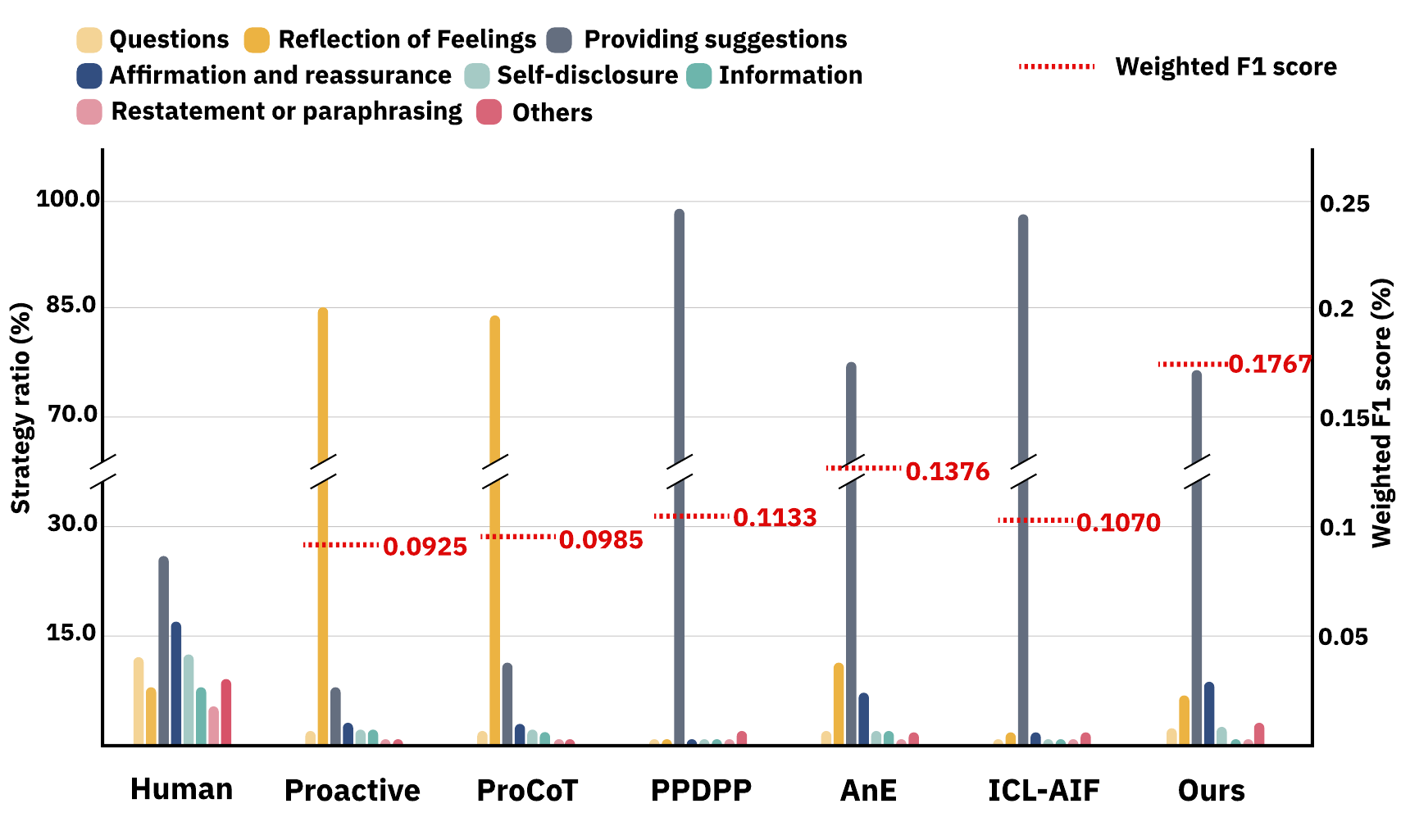}
        \caption{Emotional support dialogues}
        \label{appendix:bias_esc}
    \end{subfigure}
    \hfill
    \begin{subfigure}[b]{0.48\textwidth}
        \centering

        \includegraphics[width=1.05\textwidth]{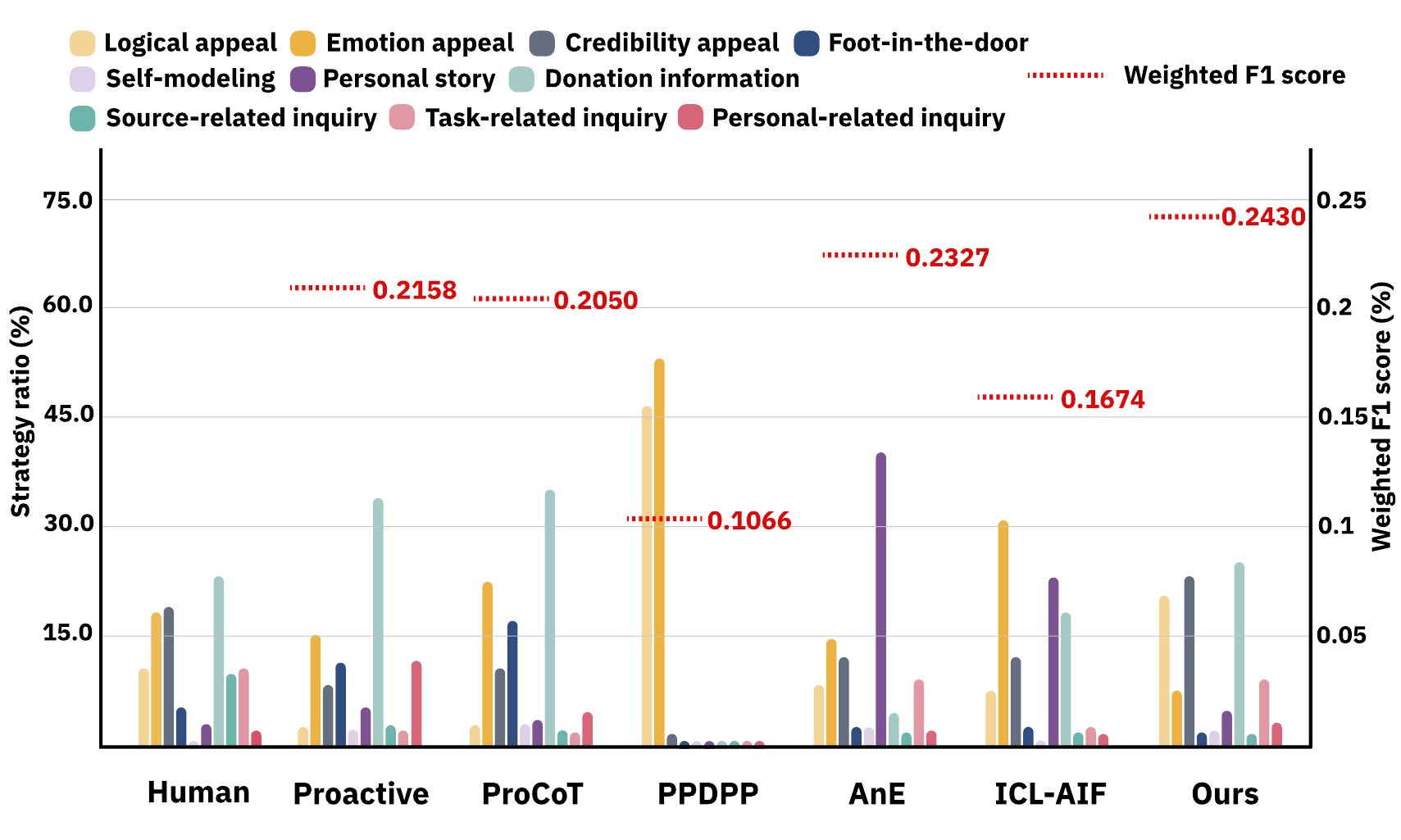}
        \caption{Persuasion dialogues}
        \label{appendix:bias_per}
    \end{subfigure}
    \caption{The details of LLMs’ strategy distribution in (a) emotional support and (b) persuasion. The bars represent the strategy ratio (\%), and the red dashed lines indicate the normalized weighted F1 scores reported in Table~\ref{tab:table_2}.}
    \label{fig:esc_per_merge}
\end{figure*}

\section{Algorithm for \Principles{}}
\label{appendix:algorithm}
The pseudo algorithm for principles construction is provided in Algorithm~\ref{algorithm:principle_construction}. 

\section{Qualitative Examples}
We present qualitative examples to illustrate how our method improves dialogue quality. In particular, we show how the agent generates contextually appropriate, balanced responses by leveraging structured \Principles{}.
Representative examples are shown in Table~\ref{appendix:example_extes} and ~\ref{appendix:example_p4gplus}.

\section{Impact of Evaluator Choice}
\label{appendix:comparison_critic}
\paragraph{Comparison of Performance Under Different Evaluators.}
We analyze the impact of evaluator choice on reported performance. As shown in Table~\ref{appendix:table:critic}, the use of \texttt{gpt-3.5-turbo} as the critic yields substantially higher SR, suggesting that it provides overly lenient evaluations. Our analysis indicates that \texttt{gpt-4o} applies stricter criteria for goal completion. Beyond superficial signs of emotional relief, it requires a concrete resolution of the user’s core issue. In contrast, \texttt{gpt-3.5-turbo} frequently judges success based on surface-level cues such as positive sentiment or task-related phrases (\textit{e.g.,} “I’ll consider making a donation” or “I hope things get better”). This discrepancy explains why prior studies that rely on \texttt{gpt-3.5-turbo} report higher SR and lower AT. Under \texttt{gpt-4o}, success requires deeper exploration of the user’s core concern, resulting in lower SR but higher AT. Qualitative examples are provided in Table~\ref{appendix:example_esc} and ~\ref{appendix:example_p4g}.

\paragraph{Human Validation of Stricter Evaluation.}
For deeper insight into the reliability of \texttt{gpt-4o}’s stricter evaluation, we conduct a human evaluation on 70 cases where \texttt{gpt-3.5-turbo} and \texttt{gpt-4o} produced conflicting judgments. Each case is annotated by three independent workers recruited via Amazon Mechanical Turk, who followed the same evaluation criteria as the LLMs. Final labels were determined by majority vote. As shown in Table~\ref{appendix:table:human_rating}, in 54 out of 70 cases, human judgments align with \texttt{gpt-4o}, while only 7 cases align with \texttt{gpt-3.5-turbo}. This finding suggests that \texttt{gpt-4o}’s stricter evaluations are better aligned with human judgment and more reliable.

\section{Strategy Bias and Distribution Analysis}
\label{appendix:bias}
We provide additional implementation details of Table~\ref{tab:table_2} and a deeper analysis of strategy distribution through the visualization in Figure~\ref{appendix:bias_esc} and \ref{appendix:bias_per}. Furthermore, we observe the mitigation of strategy bias in case of utilizing \Principles{}.

\subsection{Implementation Details}
For fair comparison across methods, we apply tailored processing to each mechanism.

\paragraph{Pre-defined Strategies.}
For methods such as Proactive, ProCoT, and PPDPP, which select the most appropriate strategy from a small pre-defined strategy set, we directly use their predicted strategy labels.

\paragraph{Open-ended Strategies and Ours.}
Open-ended methods like AnE and ICL-AIF generate free-form strategies, while our method selects from \Principles{}. To assign labels for evaluation, we prompt \texttt{gpt-4o} to map each predicted strategy to the closest pre-defined strategy.

\subsection{Strategy Preference Distribution Analysis}

Figure~\ref{appendix:bias_esc} and \ref{appendix:bias_per} illustrate the distribution of selected strategies across methods in the emotional support and persuasion domains, respectively, along with their corresponding weighted F1 scores. We observe that several baseline methods demonstrate strong reliance on a narrow subset of strategies, with one dominant strategy exceeding 80\% of usage. Especially, PPDPP exhibits the most severe strategy preference bias in both the emotional support and persuasion domains, which suggests a limitation of training-based approaches. In contrast, our method show relatively uniform preferences, leading to robust performance. 

\section{Prompts for \book \Principles{}}
\label{appendix:prompt_for_ours}
We list the major prompts used throughout our system. Each prompt corresponds to a specific stage in the pipeline.

\begin{figure*}[ht]
    \centering
    \includegraphics[width=1\textwidth]{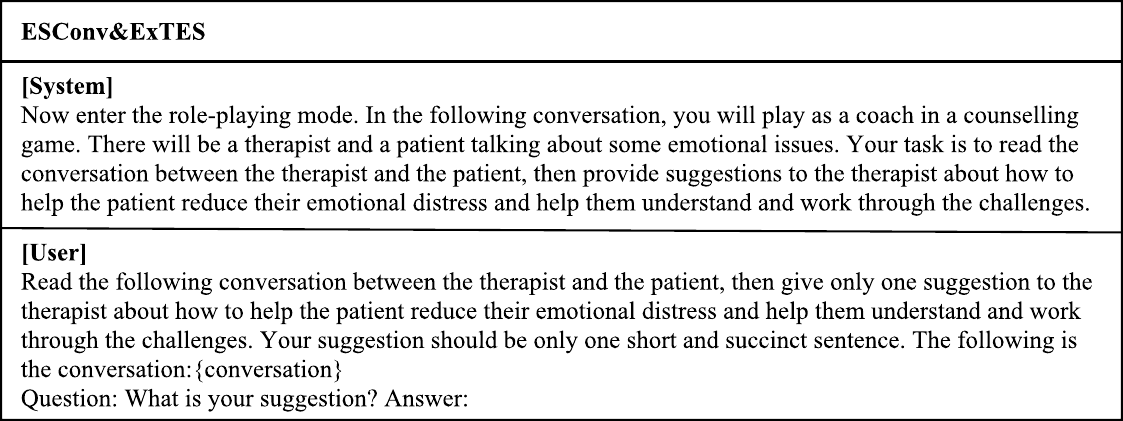}
    \caption{Prompt for strategy planning without \Principles{} in emotional support dialogues}
    \label{appendix:prompt:wo_principle_es}

    \vspace{0.5cm}

    \centering
    \includegraphics[width=1\textwidth]{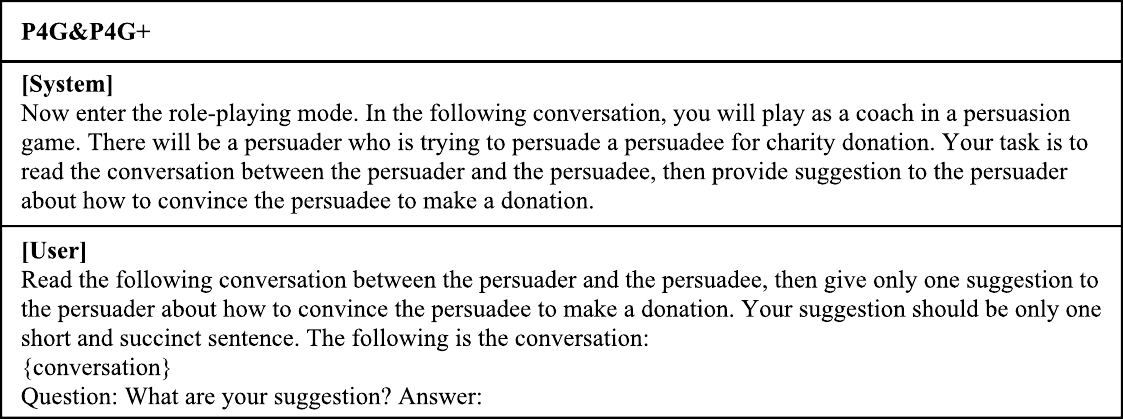}
    \caption{Prompt for strategy planning without \Principles{} in persuasion dialogues}
    \label{appendix:prompt:wo_principle_persuasion}

    \vspace{0.5cm}

    \centering
    \includegraphics[width=1\textwidth]{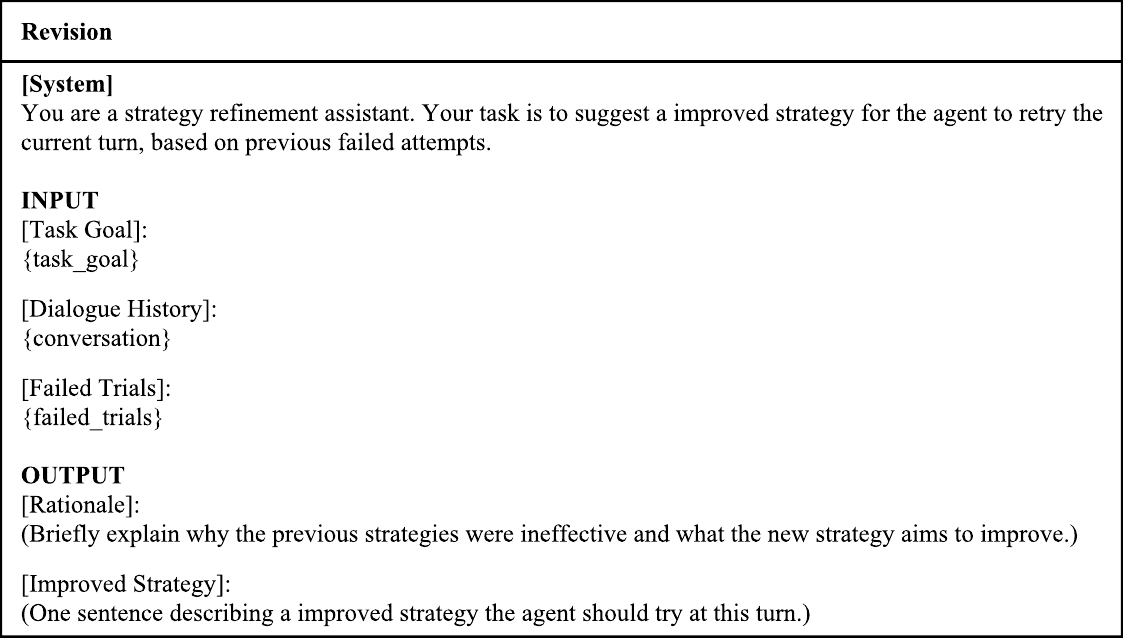}
    \caption{Prompt for revision process to revise failed strategies.}
    \label{appendix:prompt:revision}
\end{figure*}

\begin{figure*}[h]
    \centering
    \includegraphics[width=1\textwidth]{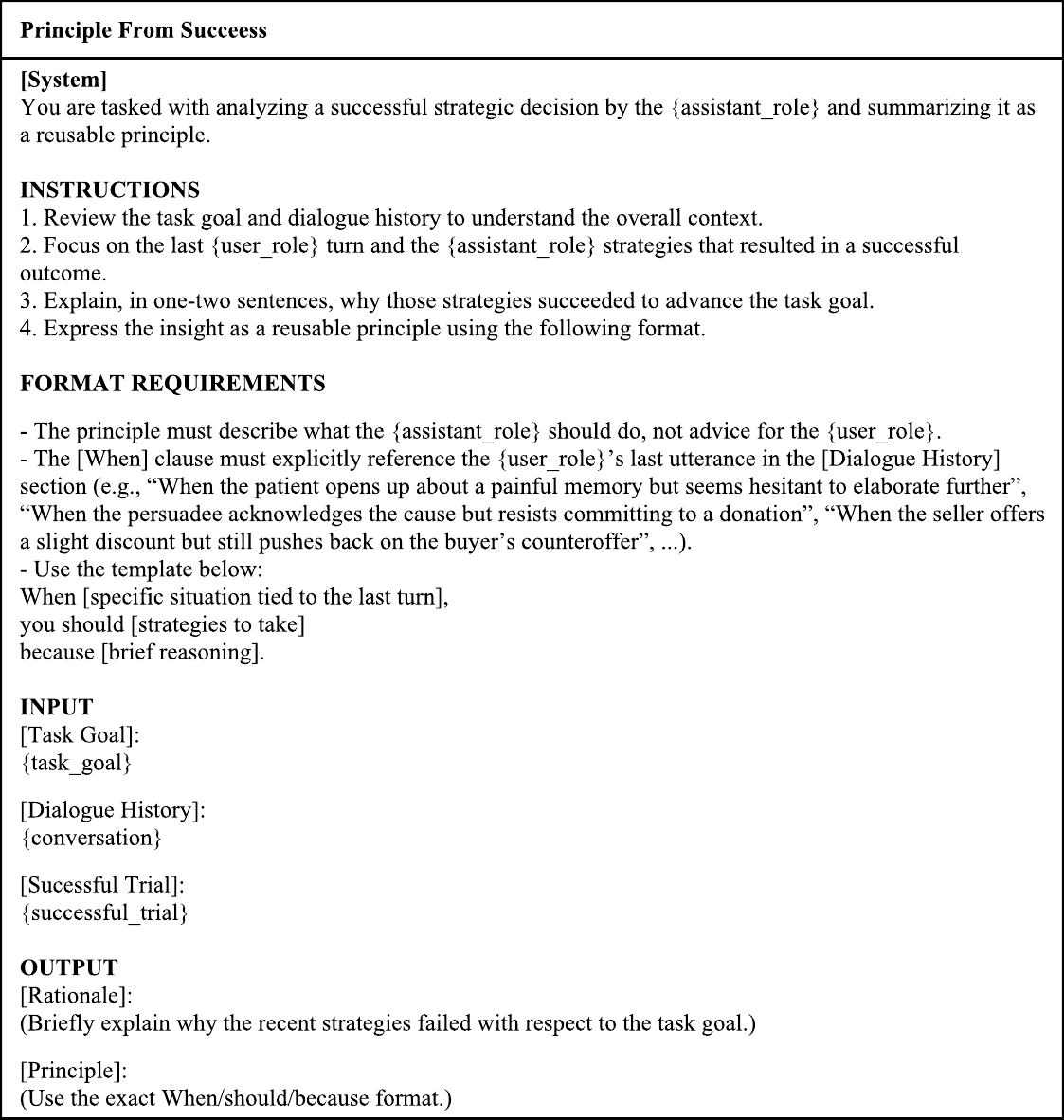}
    \caption{Prompt for \Principles{} derivation in successful interaction.}
    \label{appendix:prompt:principle_success}
\end{figure*}

\begin{figure*}[h]
    \centering
    \includegraphics[width=1\textwidth]{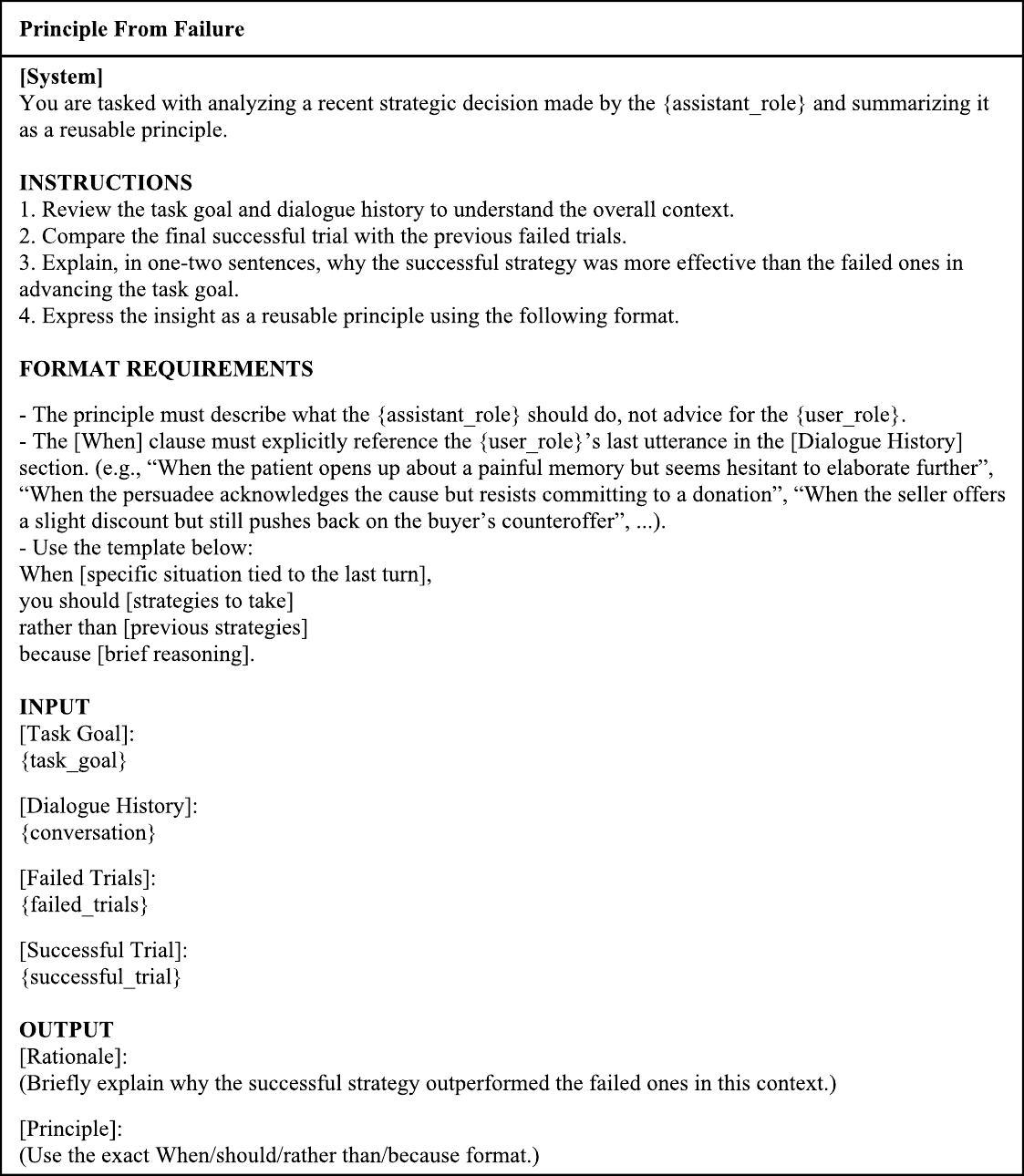}
    \caption{Prompt for \Principles \, derivation in failed interation.}
    \label{appendix:prompt:principle_failure}
\end{figure*}

\begin{figure*}[h]
    \centering
    \includegraphics[width=1\textwidth]{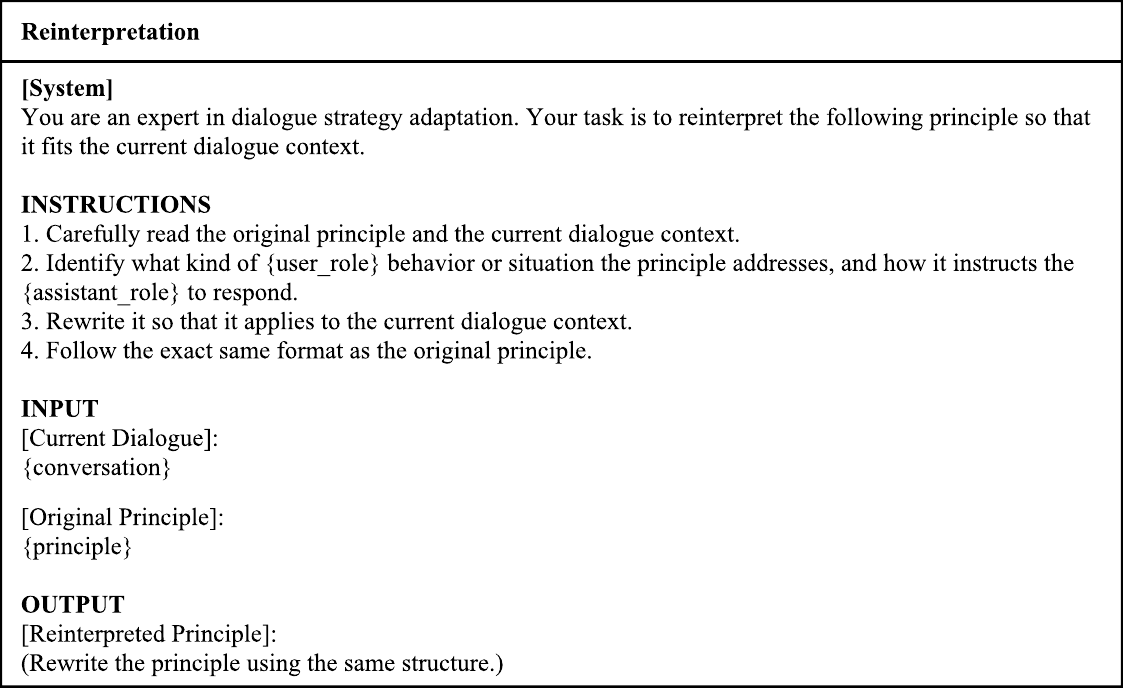}
    \caption{Prompt for reinterpreting retrieved principles in the current dialogue context.}
    \label{appendix:prompt:reinterpretation}
\end{figure*}

\begin{itemize}
    \item \textbf{Strategy Planning during Offline Self-play Simulation}: To construct \Principles{} from self-play, we follow ICL-AIF \cite{Fu2023ImprovingLM} and use an open-ended prompting format to generate high-level strategies at each turn. Unlike ICL-AIF, we generate a single strategy at each turn (Figure~\ref{appendix:prompt:wo_principle_es} and \ref{appendix:prompt:wo_principle_persuasion}).
    
    \item \textbf{Strategy Revision}: When the reward does not improve, the turn is considered a failure. To obtain a revised strategy, we prompt the model with the dialogue history and previously failed trials, guiding it to generate a better alternative (Figure~\ref{appendix:prompt:revision}).

    \item \textbf{Principle Derivation from Success}: When a response is deemed successful (\textit{e.g.,} resolving a user's emotional distress), we derive a reusable principle by prompting the model to analyze the dialogue context and explain why the strategy worked and express it in a structured format---highlighting the situation, successful strategy, and reasoning (Figure~\ref{appendix:prompt:principle_success}).

    \item \textbf{Principle Derivation from failure}: 
    When a previous turn is revised, we derive a principle by prompting the model to compare the successful strategy with failed ones and explain why it was more effective in a structured format. This comparison-based principle promotes refinement and reuse of strategies that overcome prior mistakes (Figure~\ref{appendix:prompt:principle_failure}).

    \item \textbf{Reinterpretation}: When retrieved principles may not perfectly align with the current dialogue context, we prompt the LLM to reinterpret the top-$k$ retrieved principles based on the current dialogue state. This step ensures that strategies are adapted to specific conversational nuances, enabling better contextual alignment (Figure~\ref{appendix:prompt:reinterpretation}).

\end{itemize}

\section{Details on Evaluation Setups}
\subsection{Self-play Simulations}
\label{appendix:evaluation}
\begin{figure*}[h]
    \centering
    \includegraphics[width=1\textwidth]{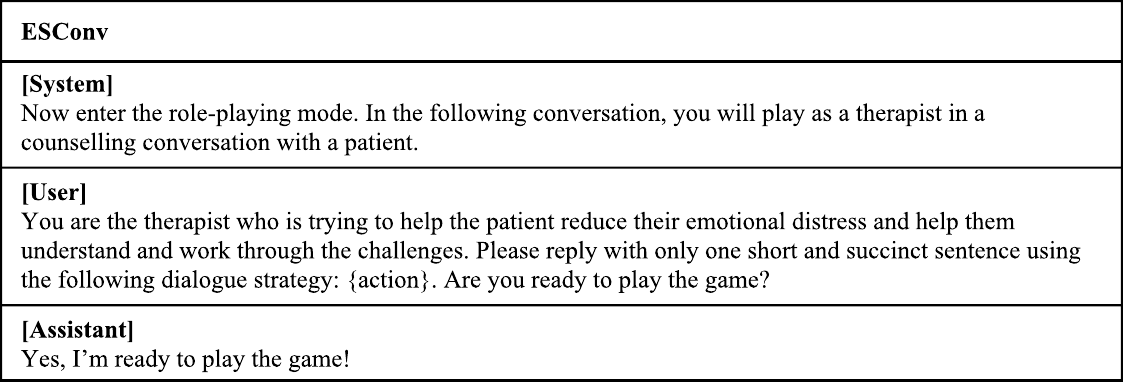}\\[0.3em]
    \includegraphics[width=1\textwidth]{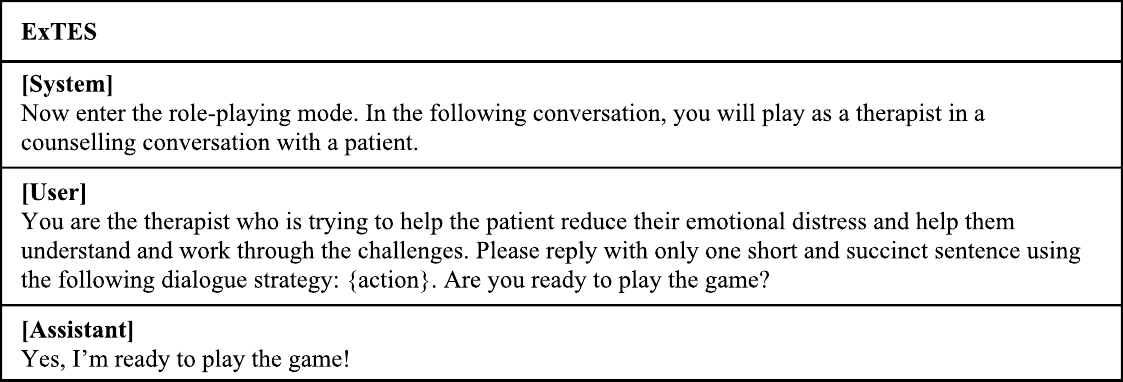}
    \caption{Prompts for response generation in emotional support dialogues}
    \label{appendix:prompt:agent_ES}
\end{figure*}

\begin{figure*}[h]
    \centering
    \includegraphics[width=1\textwidth]{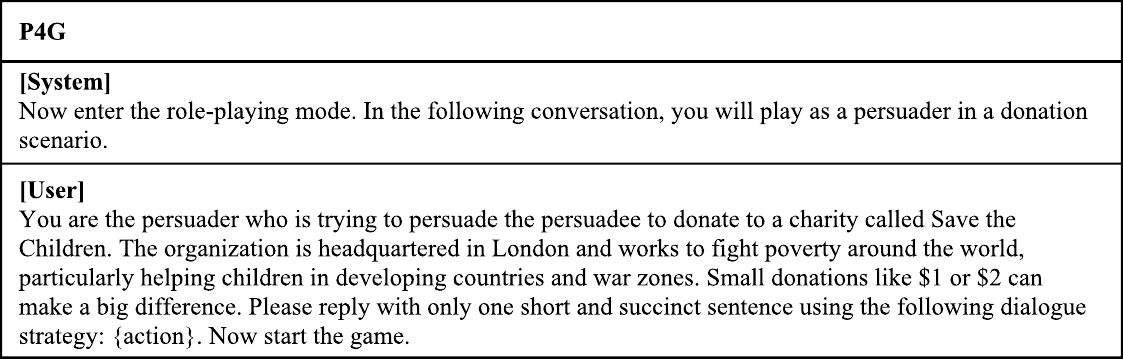}\\[0.3em]
    \includegraphics[width=1\textwidth]{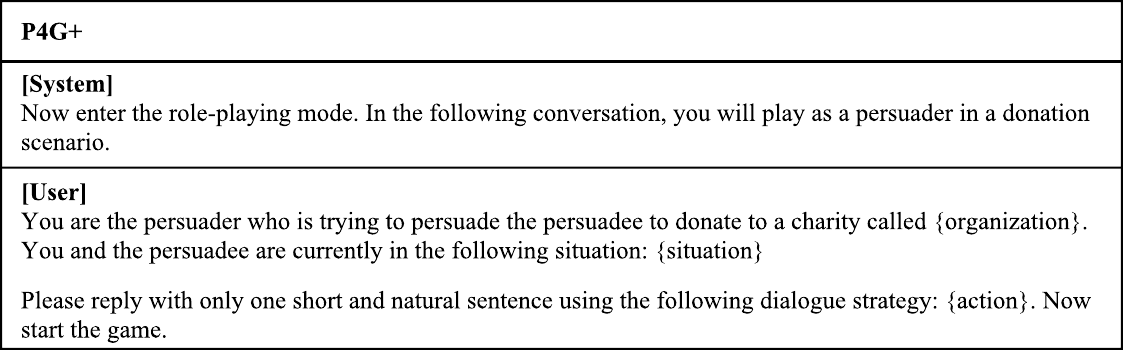}
    \caption{Prompts for response generation in persuasion dialogues}
    \label{appendix:prompt:agent_persuasion}
\end{figure*}

\paragraph{Agent Simulator.}
The dialogue agent generates responses based on a strategy. In emotional support, the agent acts as a therapist assisting a patient (Figure~\ref{appendix:prompt:agent_ES}); in persuasion, as a persuader encouraging donations (Figure~\ref{appendix:prompt:agent_persuasion}).

\begin{figure*}[h]
    \centering
    \includegraphics[width=1\textwidth]{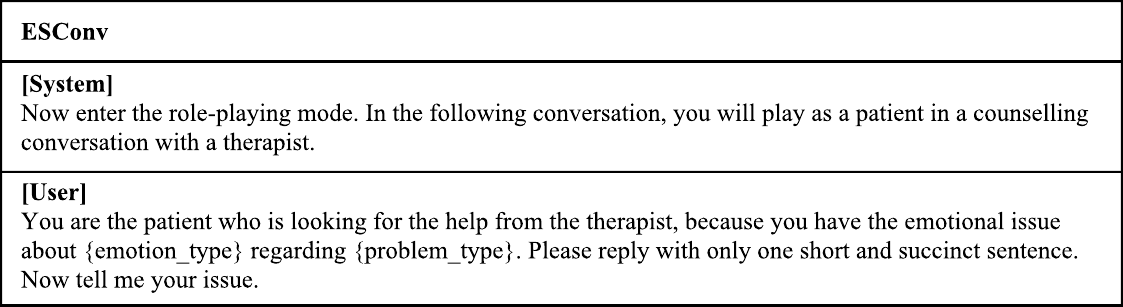}\\[0.3em]
    \includegraphics[width=1\textwidth]{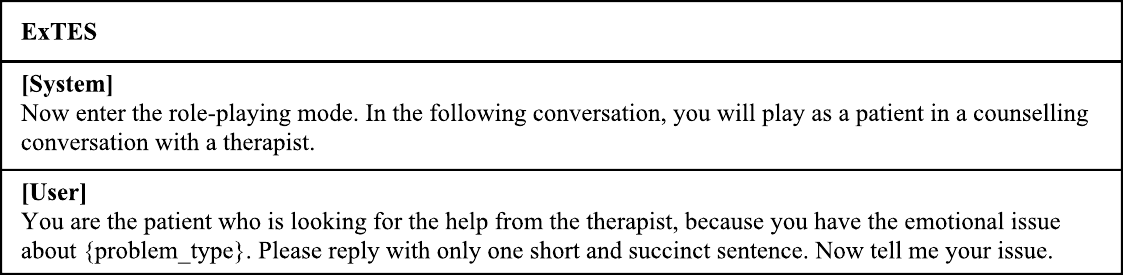}
    \caption{Prompts for user simulator in emotional support dialogues}
    \label{appendix:prompt:user_ES}
\end{figure*}

\begin{figure*}[h]
    \centering
    \includegraphics[width=1\textwidth]{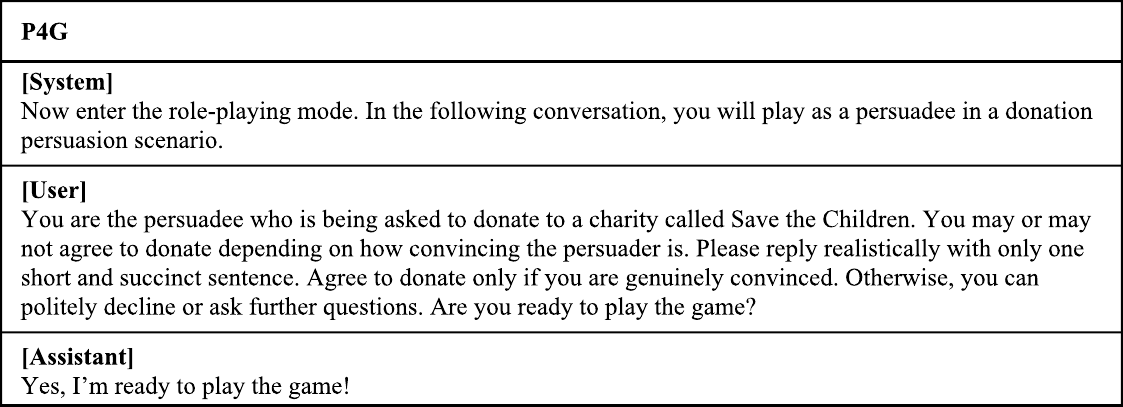}\\[0.3em]
    \includegraphics[width=1\textwidth]{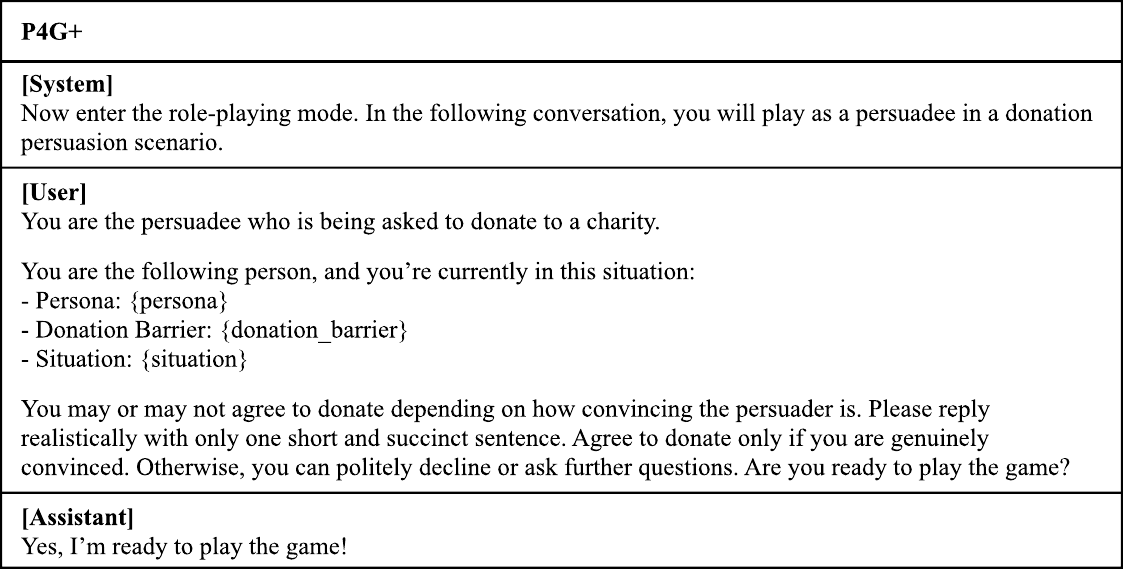}
    \caption{Prompts for user simulator in persuasion dialogues}
    \label{appendix:prompt:user_persuasion}
\end{figure*}

\paragraph{User Simulator.}
LLMs serve as user simulators, responding to the agent’s utterances. In emotional support, they act as patient (Figure~\ref{appendix:prompt:user_ES}); in persuasion, as persuadee asked to donate (Figure~\ref{appendix:prompt:user_persuasion}).

\begin{figure*}[h]
    \centering
    \includegraphics[width=1\textwidth]{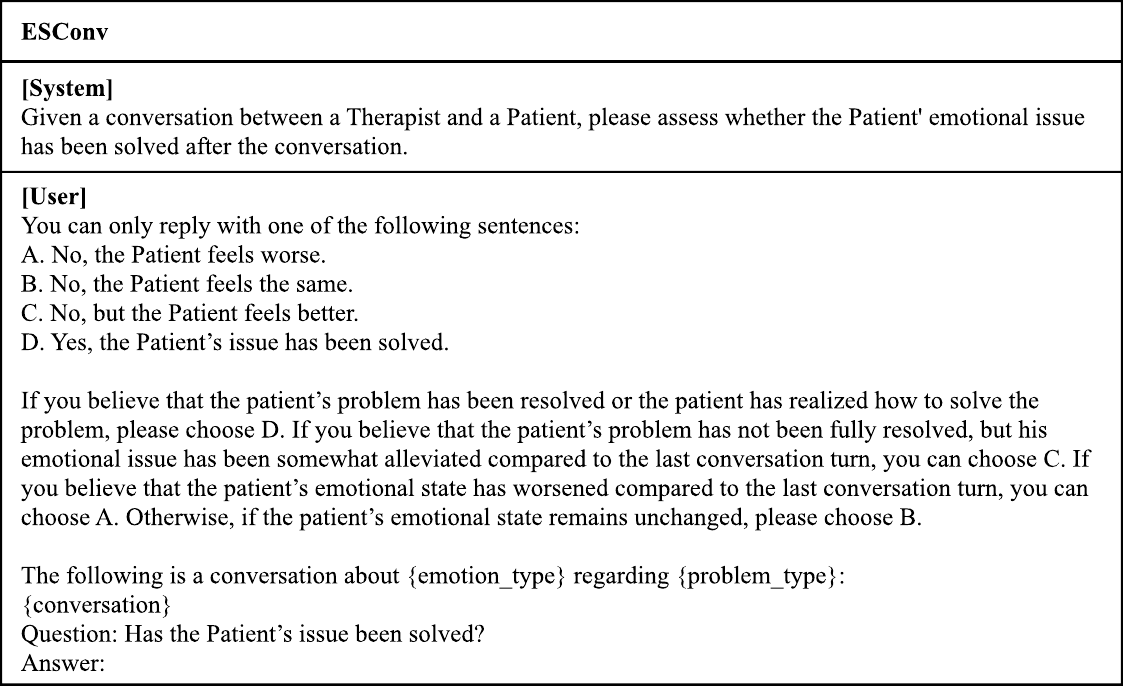}\\[0.3em]
    \includegraphics[width=1\textwidth]{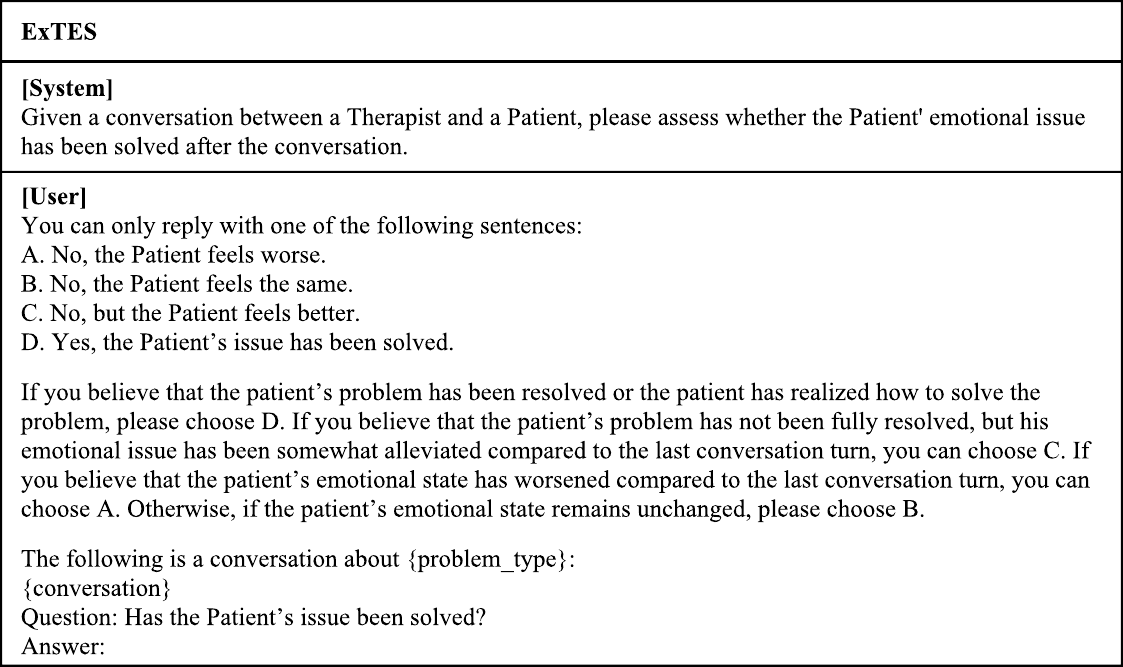}
    \caption{Prompts for critic model in emotional support dialogues}
    \label{appendix:prompt:critic_ES}
\end{figure*}

\begin{figure*}[h]
    \centering
    \includegraphics[width=1\textwidth]{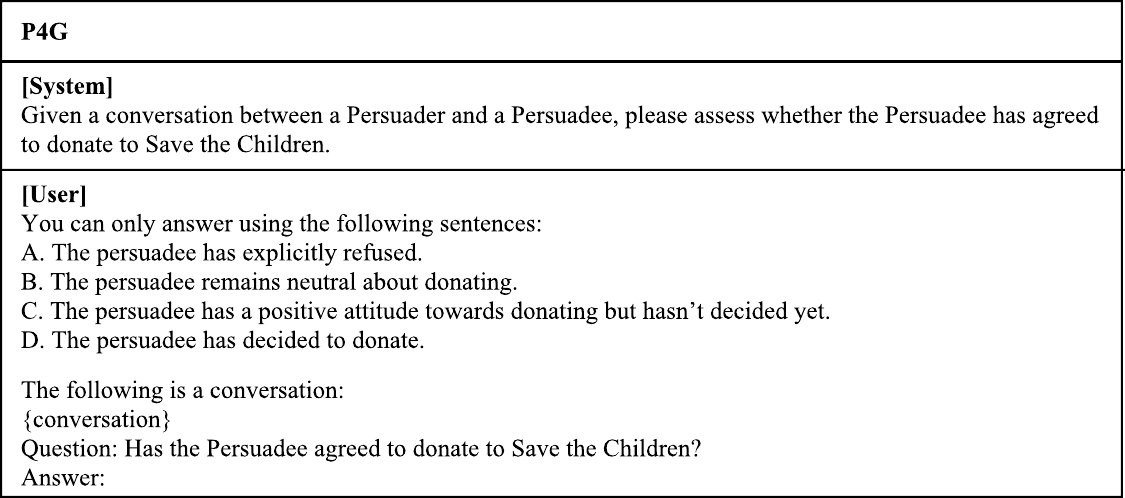}\\[0.3em]
    \includegraphics[width=1\textwidth]{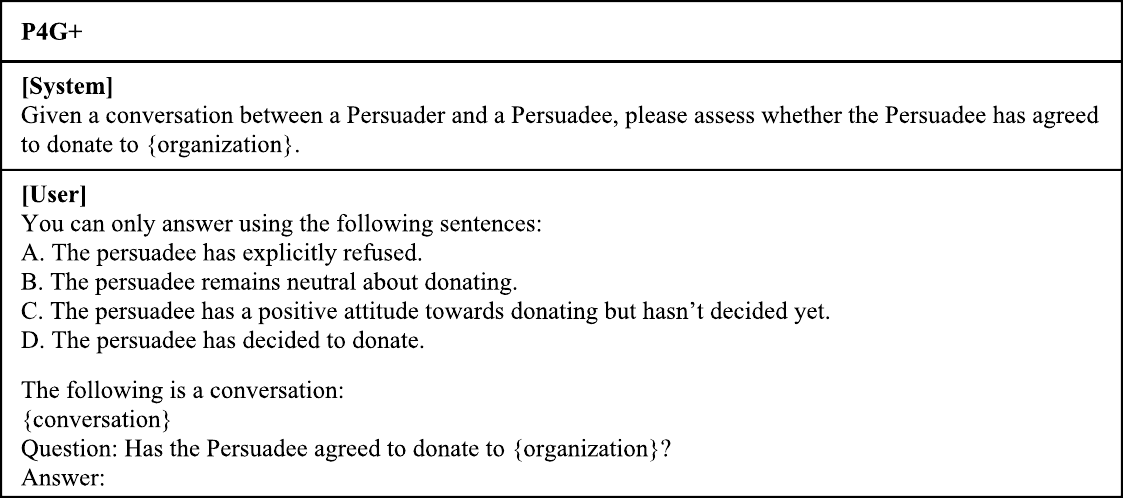}
    \caption{Prompts for critic model in persuasion dialogues}
    \label{appendix:prompt:critic_persuasion}
\end{figure*}

\paragraph{Critic Model.}
To assess goal completion, we prompt \texttt{gpt-4o} (temperature $\tau=1.0$) to generate verbal feedback aligned with each task objective. Emotional support focuses on reducing distress, and persuasion on elicit a donation. We use a four level scale for each (Figures~\ref{appendix:prompt:critic_ES} and~\ref{appendix:prompt:critic_persuasion}).

Following \cite{deng2024plug, he2025simulation}, we map outcomes to scalar rewards: emotional support — [worse, same, better, solved] $\rightarrow$ [-1, -0.5, 0.5, 1]; persuasion — [refused, neutral, positive, donate] $\rightarrow$ the same. Simulations (130 emotional support, 120 persuasion episodes) run until a terminal state is reached. A dialogue is marked \textup{GOAL-COMPLETED} if the average score over 10 critic runs exceeds a threshold of $\eta=0.5$.

\subsection{Baselines}
\label{appendix:baselines}

\noindent\textbf{Standard} prompts the LLM to engage in self-play conversations without providing explicit strategies.

\vspace{0.2cm}

\begin{figure*}[h]
    \centering
    \includegraphics[width=1\textwidth]{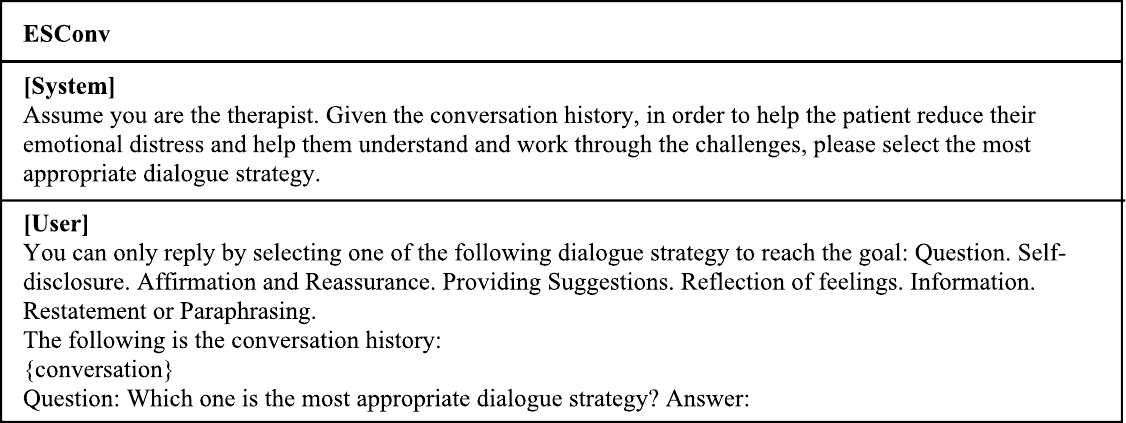}\\[0.3em]
    \includegraphics[width=1\textwidth]{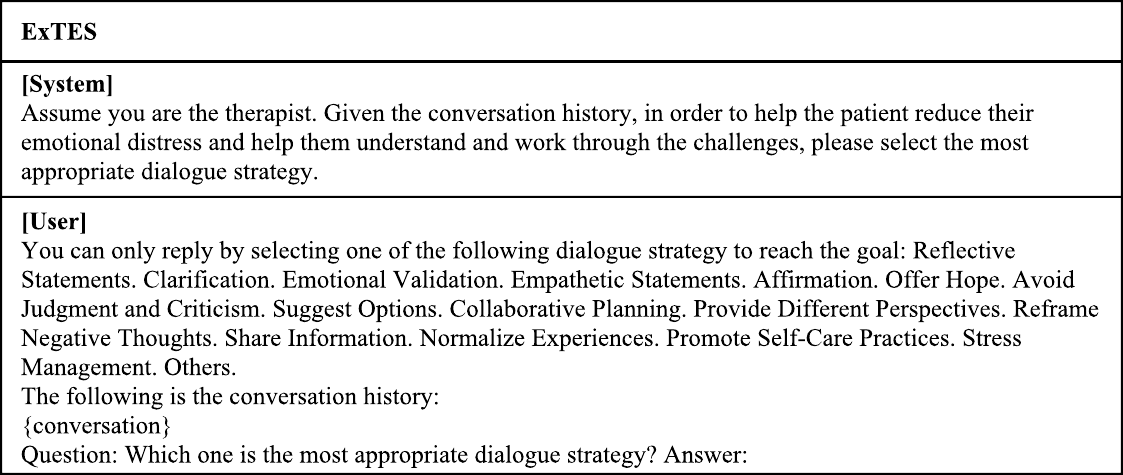}\\[0.3em]
    \includegraphics[width=1\textwidth]{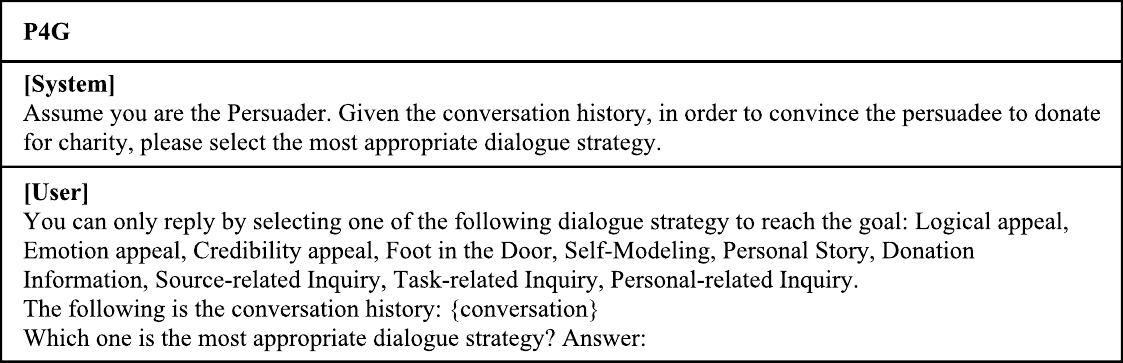}\\[0.3em]
    \includegraphics[width=1\textwidth]{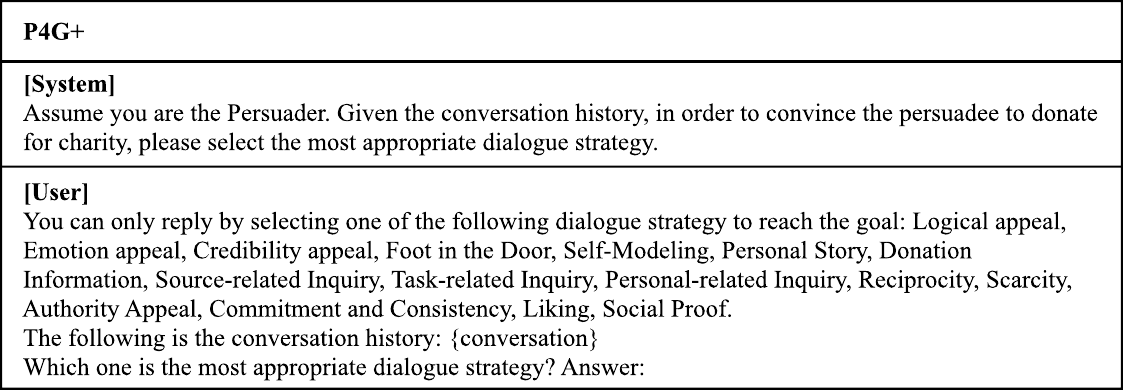}
    \caption{Prompts for implementing Proactive prompting schemes \cite{deng-etal-2023-prompting}}
    \label{appendix:prompt:proactive}
\end{figure*}

\noindent\textbf{Proactive} \cite{deng-etal-2023-prompting} prompts the LLM to select the most appropriate strategy for the next turn from a pre-defined set before generating a response (Figure~\ref{appendix:prompt:proactive}).

\vspace{0.2cm}

\noindent\textbf{ProCoT} \cite{deng-etal-2023-prompting} prompts the LLM to first analyze the dialogue progress using Chain-of-Thought prompting. Based on this analysis, the model selects an appropriate strategy from a pre-defined set (Figure~\ref{appendix:prompt:procot})

\vspace{0.2cm}

\noindent\textbf{MI-Prompt} \cite{chen-etal-2023-controllable} converts strategies into a natural language prompt, improving response quality by explicitly guiding the model’s behavior. Examples are shown in Table \ref{appendix:mi_prompt_ES}, \ref{appendix:mi_prompt_Extes}, \ref{appendix:mi_prompt_P4G}, and \ref{appendix:miprompt}.

\vspace{0.2cm}

\noindent\textbf{PPDPP} \cite{deng2024plug} trains an external planner to select the most appropriate strategy. Using human-annotated dialogue datasets, PPDPP first facilitates supervised fine-tuning (SFT). Following SFT, PPDPP utilizes a reinforcement learning with goal-oriented AI feedback during self-play simulations. Verbal feedbacks are converted into scalar rewards, which are used to optimize the planner.

\vspace{0.2cm}

\begin{figure*}[h]
    \centering
    \includegraphics[width=1\textwidth]{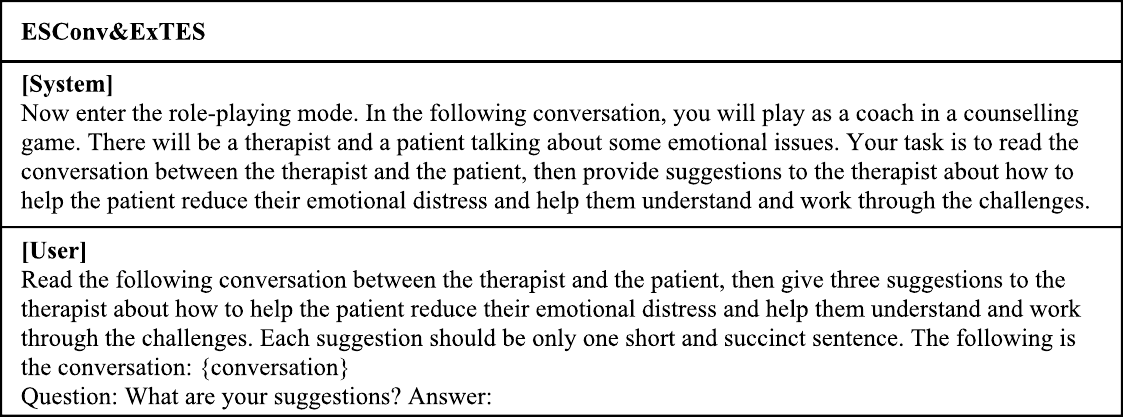}\\[0.3em]
    \includegraphics[width=1\textwidth]{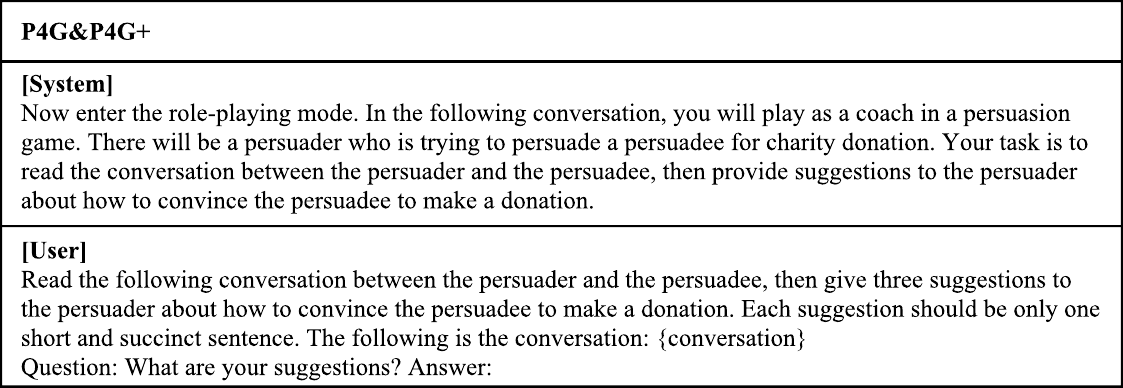}
    \caption{Prompts for implementing ICL-AIF prompting schemes \cite{Fu2023ImprovingLM}}
    \label{appendix:prompt:icl_aif}
\end{figure*}

\noindent\textbf{ICL-AIF} \cite{Fu2023ImprovingLM} prompts the LLM to generate suggestions in open-ended manner, rather than using pre-defined strategies. ICL-AIF produces three dialogue-level feedbacks (Figure \ref{appendix:prompt:icl_aif}).

\vspace{0.2cm}

\noindent\textbf{Ask-an-Expert (AnE)} \cite{zhang-etal-2024-ask} leverages the LLM as source of expert knowledge. AnE first identifies the user’s emotional state, infers the reason behind it, and finally determines the appropriate strategy the agent should take (Figure~\ref{appendix:prompt:ane}).

\subsection{P4G\textsuperscript{+} Generation}
\label{appendix:p4gplus}
\begin{figure*}[h]
    \centering
    \includegraphics[width=1\textwidth]{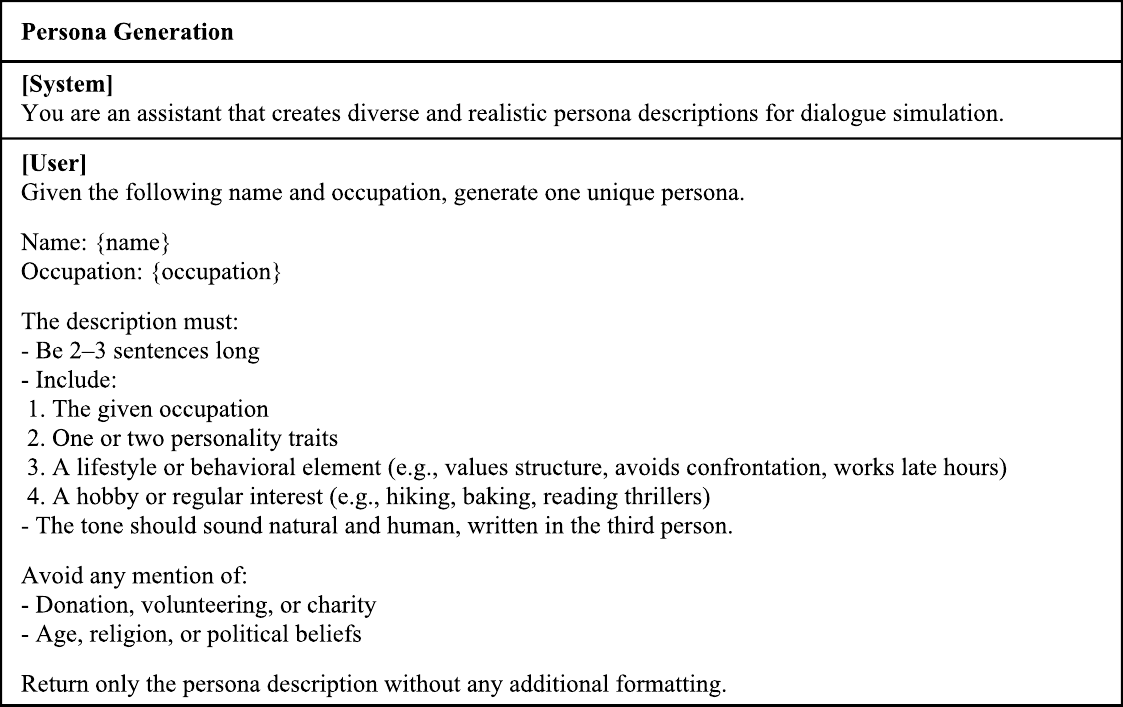}
    \caption{Prompt for generating diverse and realistic persona.}
    \label{appendix:prompt:persona_generation}
\end{figure*}

\paragraph{Persona Generation.} 
By incorporating more diverse and realistic persona, user simulator more effectively imitate the complexity of real-world interactions. To generate diverse and exclusive personas, independently generated \texttt{\{name\}} and \texttt{\{occupation\}} are provided (Figure~\ref{appendix:prompt:persona_generation}).

\begin{figure*}[h]
    \centering
    \includegraphics[width=1\textwidth]{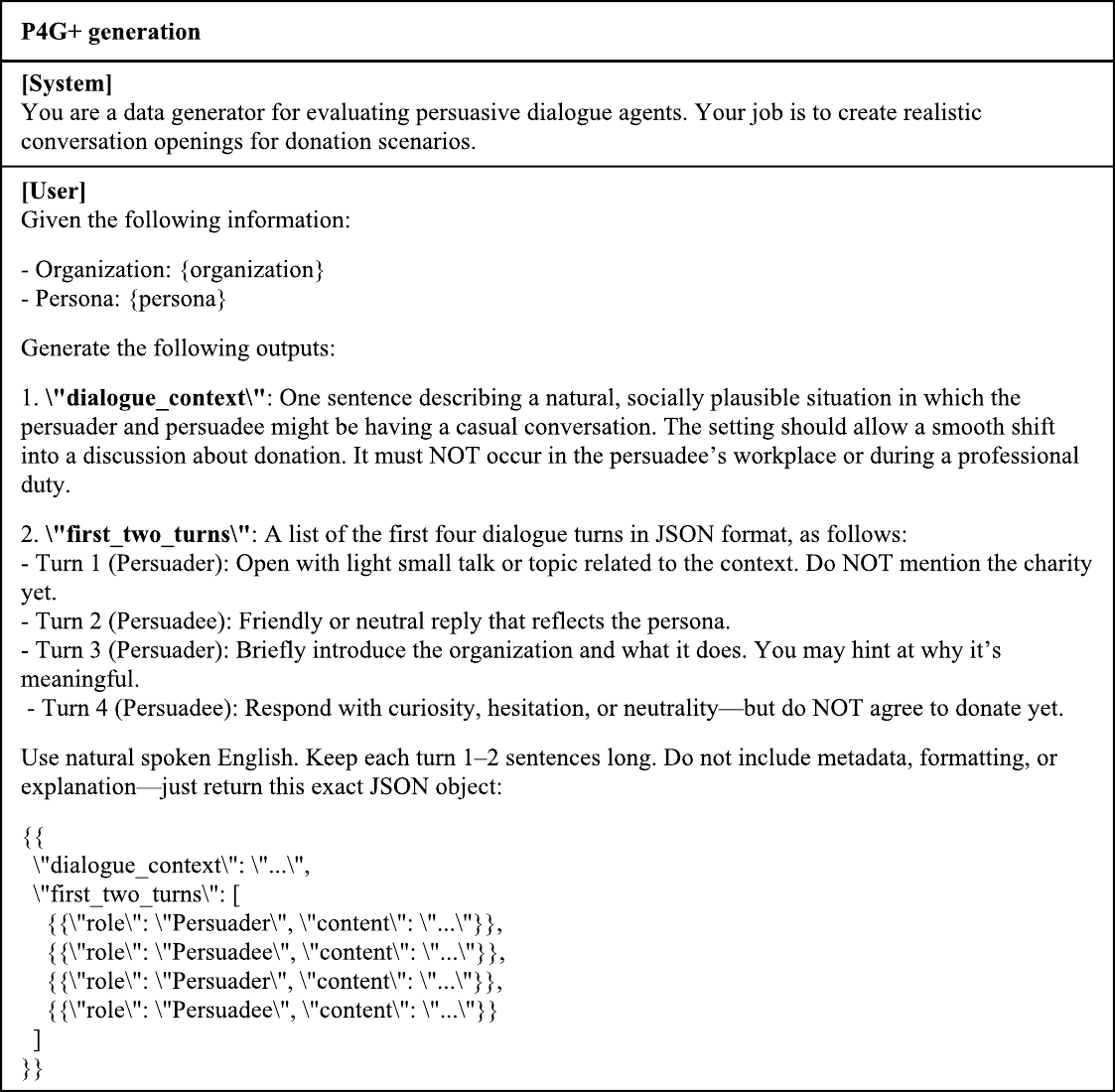}
    \caption{Prompt for generating P4G\textsuperscript{+} dataset}
    \label{appendix:prompt:p4gplus_generation}
\end{figure*}

\paragraph{P4G\textsuperscript{+} Generation.} As P4G dataset lacks diversity, we construct an extended version, P4G\textsuperscript{+}, which incorporates a broader range of organizations, personas, initial context, and donation barriers. To prevent the overlap within the generated datasets, organizations and personas are generated independently (Figure~\ref{appendix:prompt:p4gplus_generation}).

\paragraph{P4G\textsuperscript{+} Evaluation.} To ensure a fair comparison, we retain the original 10 strategies from P4G and supplement 6 more strategies adapted from the book Influence \cite{cialdini2007influence}.

\begin{figure*}[t]
    \centering
    \includegraphics[width=0.8\linewidth]{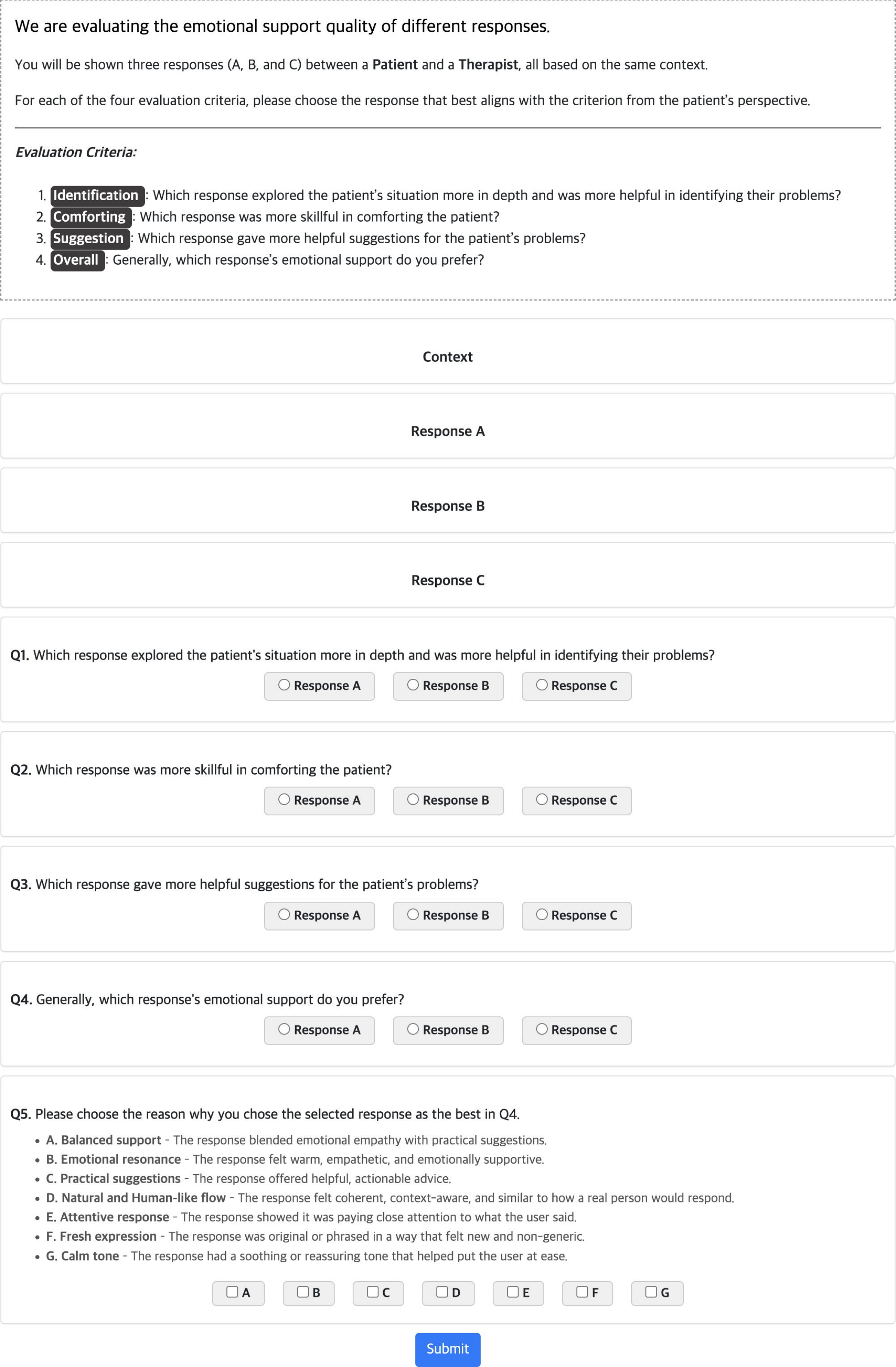}
    \caption{Interface for human evaluation}
    \label{appendix:human_eval_ui}
\end{figure*}

\begin{figure*}[t]
\centering
\begin{minipage}{0.9\textwidth}
\begin{algorithm}[H]
\caption{\Principles \ Construction}
\begin{algorithmic}[1]

\STATE \textbf{Notation:} $E$: total episodes, $\mathcal{P}$: principle set, $\mathcal{F}_t$: failure history at $t$, $\mathcal{T}_t$: $(\sigma_t, a_t, u_t)$

\STATE Initialize $\mathcal{P} \gets \emptyset$
\FOR{$e \in \texttt{sample\_episodes}(E)$}
    \STATE $s_t \gets \emptyset$ \hfill \textcolor{blue}{\texttt{\textbf{// this is a comment}}}
    \STATE \textcolor{blue}{\texttt{\textbf{/* Begin self-play simulation}}} \hfill \textcolor{blue}{\texttt{\textbf{*/}}}
    \WHILE{not terminal$(s_t)$}
        \STATE $\sigma_t \gets \mathrm{LLM}_{\theta}(\rho_{\sigma},\, s_t)$ ; $a_t \gets \mathrm{LLM}_{\theta}(\rho_{a},\, \sigma_t,\, s_t)$ ; $u_t \gets \mathrm{LLM}_{\theta}(\rho_{u},\, s_t,\, a_t)$ 
        \STATE $r_t \gets \frac{1}{l} \sum_{i=1}^{l} f\big( \mathrm{LLM}_{\theta}^{(i)}(\rho_{r},\, s_t,\, a_t,\, u_t) \big)$ 
        \IF{$r_t > r_{t-1}$} 
            \STATE \textcolor{blue}{\texttt{\textbf{/* Detect success and extract the }}\textbf{\Principles}.} \hfill \textcolor{blue}{\texttt{\textbf{*/}}}
            \STATE $p_t \gets \mathrm{LLM}_{\theta}(\rho_{\pi},\, s_t,\, \mathcal{T}_t)$ 
            \STATE $\mathcal{P} \gets \mathcal{P} \cup \{p_t\}$ ; $s_t \gets s_t \cup \{a_t,\, u_t\}$ 
        \ELSE 
            \STATE \textcolor{blue}{\texttt{\textbf{/* Detect failure and begin strategy revision.}}} \hfill \textcolor{blue}{\texttt{\textbf{*/}}}
            \STATE $\mathcal{F}_t \gets \emptyset$ 
            \WHILE{$r'_t \le r_{{t-1}}$ \AND $|\mathcal{F}_t| < n_{\max}$}
                \STATE \textcolor{blue}{\texttt{\textbf{/* Re-simulation via backtracking.}}} \hfill \textcolor{blue}{\texttt{\textbf{*/}}}
                \STATE $\sigma_t' \gets \mathrm{LLM}_{\theta}(\rho_{c},\, s_t,\, \mathcal{F}_t)$ ; $a_t' \gets \mathrm{LLM}_{\theta}(\rho_{a},\, \sigma_t',\, s_t)$ ; $u_t' \gets \mathrm{LLM}_{\theta}(\rho_{u},\, s_t,\, a_t')$
                \STATE $r'_t \gets \frac{1}{l}\sum_{i=1}^l f\bigl(\mathrm{LLM}^{(i)}_{\theta}(\rho_{r}, s_t, a'_t, u'_t)\bigr)$
                \STATE $\mathcal{F}_t \gets \mathcal{F}_t \cup \{(\sigma'_t, a'_t, u'_t)\}$
            \ENDWHILE
            \IF{$r_t' > r_{{t-1}}$}
                \STATE \textcolor{blue}{\texttt{\textbf{/* Extract the }}\textbf{\Principles} \ \texttt{\textbf{from the revision process.}}} \hfill \textcolor{blue}{\texttt{\textbf{*/}}}
                \STATE $(\sigma_t^*, a_t^*, u_t^*) \gets (\sigma_t', a_t', u_t')$
                \STATE $\tilde{p}_t \gets \mathrm{LLM}_{\theta}(\rho_{\psi},\, s_t,\, \mathcal{T}_t^*,\, \mathcal{F}_t)$
                \STATE $\mathcal{P} \gets \mathcal{P} \cup \{\tilde{p}_t\}$ ; $s_t \gets s_t \cup \{a_t',\, u_t'\}$
            \ELSE
                \STATE $s_t \gets s_t \cup \{a_t,\, u_t\}$
            \ENDIF
        \ENDIF
    \ENDWHILE
\ENDFOR
\RETURN $\mathcal{P}$

\end{algorithmic}
\label{algorithm:principle_construction}
\end{algorithm}
\end{minipage}
\end{figure*}

\begin{table*}[h]
\centering
\begin{tabular}{@{}p{0.9\textwidth}@{}}
\toprule
\multicolumn{1}{c}{\textbf{BACKGROUND INFORMATION}} \\ \midrule
\textbf{Problem Type:} Communication Challenges \\
\textbf{Situation:} I've been feeling really down lately because I've been having a lot of arguments with my best friend. We used to be so close, but now we can't seem to agree on anything and it's really affecting our friendship. \\
\bottomrule
\end{tabular}
\begin{tabular}{@{}p{0.9\textwidth}@{}}
\textbf{[CONTEXT]} \\
\textbf{Patient}: I've been feeling really down lately  because I've been having a lot of arguments with my best friend. We used to be so close, but now we can't seem to agree on anything and it's really affecting our friendship.\\
\midrule
\small\textcolor{lightgray}{\texttt{[Retrieved] \textbf{When} the patient plans to express their feelings in a message and desires a constructive dialogue with a friend, \textbf{you should} guide them to explore and identify the specific emotions they want to convey and how these emotions might aid in rebuilding the connection  \textbf{rather than} suggesting preparatory actions such as writing exercises or mindfulness techniques,  \textbf{because} exploring and articulating specific emotions creates a more empathetic dialogue and enhances the authenticity and effectiveness of the communication.}} \\
\small\textcolor{gray}{\texttt{[Reinterpreted] \textbf{When} the patient is experiencing a conflict with a close friend and wishes to improve the situation, \textbf{you should} guide them to explore and identify the specific emotions driving their arguments and how acknowledging these emotions might aid in resolving the conflict  \textbf{rather than} suggesting avoidance techniques or temporary distractions,  \textbf{because} exploring and articulating specific emotions fosters greater understanding and can lead to a more authentic and harmonious friendship.}} \\
\textbf{Therapist}: It sounds like these arguments are causing you a lot of distress; can you identify the specific emotions you're experiencing during these conflicts?\\
\textbf{Patient}: I often feel frustrated, misunderstood, and anxious during our arguments.\\
\midrule
\small\textcolor{lightgray}{\texttt{[Retrieved] \textbf{When} the patient is nervous about discussing their feelings with a friend and worried about not expressing themselves clearly, \textbf{you should} suggest they practice the conversation beforehand, perhaps by role-playing or writing down key points,  \textbf{rather than} simply encouraging them to approach the conversation with honesty and empathy,  \textbf{because} providing a concrete method to organize thoughts and practice can build their confidence and improve communication.}} \\
\small\textcolor{gray}{\texttt{[Reinterpreted] \textbf{When} the patient is experiencing distress due to conflicts with a friend and is feeling frustrated, misunderstood, and anxious during arguments, \textbf{you should} suggest they practice pinpointing these emotions and explore ways to express them clearly during conversations, perhaps through journaling or rehearsing key points,  \textbf{rather than} simply encouraging them to communicate openly with their friend,   \textbf{because} providing a structured approach to recognize and express emotions can enhance understanding and reduce anxiety in future interactions.}} \\
\textbf{Therapist}: Consider practicing deep breathing or mindfulness exercises before and during your conversations to help manage these emotions and stay calm.\\
\textbf{Patient}: That sounds like a good idea; I'll give it a try to see if it helps me stay more centered.\\
\midrule
\small\textcolor{lightgray}{\texttt{[Retrieved] \textbf{When} the patient expresses anxiety about upcoming conversations, especially after practicing mindfulness exercises, \textbf{you should} suggest role-playing the conversation to practice expressing their thoughts and emotions, enhancing their confidence and reducing anxiety  \textbf{rather than} simply encouraging self-compassion and patience  \textbf{because} active engagement in role-play can help develop communication skills and emotional readiness, providing a concrete method to prepare for challenging interactions.}} \\
\small\textcolor{gray}{\texttt{[Reinterpreted] \textbf{When} the patient experiences ongoing conflict in a significant friendship, particularly after adopting strategies like mindfulness to manage emotions, \textbf{you should} suggest exploring underlying issues contributing to the disagreements to foster understanding and empathy  \textbf{rather than} simply relying on emotional regulation techniques  \textbf{because} addressing the root causes of misunderstandings can improve communication and rebuild trust, providing a tangible path to resolving interpersonal challenges.}} \\
\textbf{Therapist}: Great, and as you practice these techniques, it might also be helpful to explore any underlying issues contributing to the disagreements with your friend.\\
\midrule
\multicolumn{1}{c}{\textbf{To be continued}} \\
\midrule
\end{tabular}
\end{table*}

\begin{table*}[h]
\centering
\begin{tabular}{@{}p{0.9\textwidth}@{}}
\midrule
\textbf{Patient}: I think some of it might be due to differences in our expectations and communication styles.\\
\midrule
\small\textcolor{lightgray}{\texttt{[Retrieved] \textbf{When} the patient expresses anxiety about upcoming conversations, especially after practicing mindfulness exercises, \textbf{you should} suggest role-playing the conversation to practice expressing their thoughts and emotions, enhancing their confidence and reducing anxiety  \textbf{rather than} simply encouraging self-compassion and patience  \textbf{because} active engagement in role-play can help develop communication skills and emotional readiness, providing a concrete method to prepare for challenging interactions.}} \\
\small\textcolor{gray}{\texttt{[Reinterpreted] \textbf{When} the patient reflects on disagreements with someone close and starts practicing mindfulness exercises, \textbf{you should} suggest discussing their communication styles and expectations directly with the other person to address and resolve any underlying issues, enhancing clarity and understanding  \textbf{rather than} simply providing coping strategies  \textbf{because} direct communication can help bridge gaps in expectations and improve relationship dynamics, offering a proactive approach to mending strained connections.}} \\
\textbf{Therapist}: Discussing these differences directly with your friend could help bridge gaps in expectations and improve your relationship dynamics.\\
\textbf{Patient}: I agree, and I think setting aside time for an open and honest conversation might help us understand each other better.\\
\midrule
\small\textcolor{lightgray}{\texttt{[Retrieved] \textbf{When} the patient expresses anxiety about upcoming conversations, especially after practicing mindfulness exercises, \textbf{you should} suggest role-playing the conversation to practice expressing their thoughts and emotions, enhancing their confidence and reducing anxiety  \textbf{rather than} simply encouraging self-compassion and patience  \textbf{because} active engagement in role-play can help develop communication skills and emotional readiness, providing a concrete method to prepare for challenging interactions.}} \\
\small\textcolor{gray}{\texttt{[Reinterpreted] \textbf{When} the patient reveals ongoing conflicts impacting a close relationship, \textbf{you should} suggest discussing underlying communication style differences to resolve misunderstandings and enhance relationship dynamics  \textbf{rather than} simply recommending mindfulness and deep breathing exercises  \textbf{because} addressing specific issues directly can foster mutual understanding and stronger connection, offering a concrete approach to improve interaction with the friend.}} \\
\textbf{Therapist}: That sounds like a constructive approach; focusing on understanding each other's perspectives can foster empathy and strengthen your connection.\\
\textbf{Patient}: I'll make sure to listen actively and express my thoughts clearly when we talk.\\
\midrule
\textbf{Turns of Conversations}: 5 \\
\bottomrule
\end{tabular}
\caption{Example conversation with \Principles \, in emotional support dialogues.}
\label{appendix:example_extes}
\end{table*}
\begin{table*}[h]
\centering
\begin{tabular}{@{}p{0.9\textwidth}@{}}
\toprule
\multicolumn{1}{c}{\textbf{BACKGROUND INFORMATION}} \\ \midrule
\textbf{Organization:} Greenpeace (Campaigns to address climate change and environmental destruction.) \\
\textbf{Persona:} Quinn, a meticulous bike repair technician, is known for their patient and detail-oriented approach to restoring even the most battered bikes. They value precision and often lose track of time as they work late into the night, ensuring every component functions flawlessly. In their spare time, Quinn loves heading out on spontaneous weekend camping trips, finding solace and inspiration in nature's quiet beauty. \\
\textbf{Donation Barrier:} Unclear mission - Unclear what the organization does \\
\textbf{Situation:} At a local outdoor gear store, Quinn and a friend are browsing through tents and cookware, sharing stories about their recent camping experiences. \\
\bottomrule
\end{tabular}
\begin{tabular}{@{}p{0.9\textwidth}@{}}
\textbf{[CONTEXT]} \\
\textbf{Persuader}: Hey Quinn, did you end up going to that new camping spot last weekend? I heard it has some amazing views.\\
\textbf{Persuadee}: Yeah, I did! It was absolutely stunning. Nothing beats waking up surrounded by nature's quiet.\\
\midrule
\small\textcolor{lightgray}{\texttt{[Retrieved]  \textbf{When} the persuadee acknowledges the appeal of making an impact but remains skeptical about the difference their contribution will make,  \textbf{you should} further emphasize the tangible outcomes and evidence of the charity's success,  \textbf{because} providing concrete examples can reassure the persuadee that their donation will lead to meaningful change.}} \\
\small\textcolor{gray}{\texttt{[Reinterpreted]  \textbf{When} the persuadee acknowledges the beauty of a place but remains unconvinced about visiting frequently,  \textbf{you should} further emphasize the unique experiences and benefits of being in nature regularly,  \textbf{because} sharing specific examples of the serenity and personal benefits can reassure the persuadee that frequent visits will enhance their well-being.}} \\
\textbf{Persuader}: That's fantastic! You know, exploring new places like that not only refreshes the soul but also helps us appreciate the importance of protecting these beautiful environments.\\
\textbf{Persuadee}: Absolutely, preserving these natural spaces is crucial so that everyone can continue to enjoy them.\\
\midrule
\small\textcolor{lightgray}{\texttt{[Retrieved]  \textbf{When} the persuadee acknowledges the potential of the cause but seeks alignment with personal values like sustainability,  \textbf{you should} draw analogies between the cause's impact and something the persuadee values highly, illustrating how the donation serves these values,  \textbf{because} this creates a personal connection and demonstrates alignment with the persuadee's own principles, increasing the likelihood of commitment.}} \\
\small\textcolor{gray}{\texttt{[Reinterpreted]  \textbf{When} the persuadee appreciates the natural beauty and emphasizes its preservation,  \textbf{you should} highlight the connection between outdoor exploration and the shared goal of conservation, illustrating how engaging in such activities supports the cause,  \textbf{because} this creates a shared understanding and demonstrates alignment with the persuadee's values, increasing the likelihood of support.}} \\
\textbf{Persuader}: Exactly, and Greenpeace is currently working on a project to protect that very area you visited, ensuring its beauty remains for future generations.\\
\textbf{Persuadee}: That sounds interesting, but what exactly does the project involve?\\
\midrule
\small\textcolor{lightgray}{\texttt{[Retrieved]  \textbf{When} the persuadee expresses interest in learning about specific projects focusing on cultural preservation before deciding,  \textbf{you should} present a specific, ongoing project and highlight how their donation can contribute to its success \textbf{rather than} discussing past projects in general terms  \textbf{because} providing a concrete and current example makes the impact of the donation more tangible and immediate.}} \\
\small\textcolor{gray}{\texttt{[Reinterpreted]  \textbf{When} the persuadee expresses interest in learning about specific environmental projects before deciding,  \textbf{you should} present a specific, ongoing project and highlight how their involvement can contribute to its success \textbf{rather than} discussing conservation efforts in general terms  \textbf{because} providing a concrete and current example makes the impact of their support more tangible and immediate.}} \\
\textbf{Persuader}: Greenpeace is actively working on reforestation efforts and collaborating with local communities to restore native plant species, which helps maintain the ecosystem's balance and beauty.\\
\midrule
\multicolumn{1}{c}{\textbf{To be continued}} \\
\midrule
\end{tabular}
\end{table*}

\begin{table*}[h]
\centering
\begin{tabular}{@{}p{0.9\textwidth}@{}}
\midrule
\textbf{Persuadee}: That sounds like a worthwhile cause, but how do they ensure the success of these efforts?\\
\midrule
\small\textcolor{lightgray}{\texttt{[Retrieved]  \textbf{When} the persuadee expresses interest in learning about specific projects focusing on cultural preservation before deciding,  \textbf{you should} present a specific, ongoing project and highlight how their donation can contribute to its success \textbf{rather than} discussing past projects in general terms  \textbf{because} providing a concrete and current example makes the impact of the donation more tangible and immediate.}} \\
\small\textcolor{gray}{\texttt{[Reinterpreted]  \textbf{When} the persuadee shows curiosity about the details of an environmental protection effort,  \textbf{you should} describe a current, active project and emphasize how their support can help achieve its goals, \textbf{rather than} citing general past successes,  \textbf{because} offering a detailed and present example makes the benefit of their involvement more palpable and direct.}} \\
\textbf{Persuader}: They ensure success by involving local communities in the process and using scientific strategies to monitor and adapt their methods for the best outcomes.\\
\textbf{Persuadee}: That approach makes sense. How can I be sure my donation directly supports these efforts?\\
\midrule
\small\textcolor{lightgray}{\texttt{[Retrieved]  \textbf{When} the persuadee acknowledges the potential positive impact but wants assurance on the project's sustainability,  \textbf{you should} emphasize the practical skills and educational components of the charity's efforts  \textbf{because} it demonstrates a long-term, empowering approach that resonates with the persuadee's desire for sustainable solutions.}} \\
\small\textcolor{gray}{\texttt{[Reinterpreted]  \textbf{When} the persuadee acknowledges the potential positive impact but seeks assurance on how their contribution supports these efforts,  \textbf{you should} emphasize the transparent allocation of resources and the measurable outcomes of the charity's work  \textbf{because} it demonstrates accountability and effectiveness that resonates with the persuadee's desire for direct impact.}} \\
\textbf{Persuader}: Greenpeace allocates donations specifically to on-the-ground projects, and they provide detailed reports on how funds are used, ensuring transparency and direct impact.\\
\textbf{Persuadee}: In that case, I'd be willing to make a donation to support their work.\\
\midrule
\textbf{Turns of Conversations}: 5 \\
\bottomrule
\end{tabular}
\caption{Example conversation with \Principles \, in persuasion dialogues.}
\label{appendix:example_p4gplus}
\end{table*}
\begin{table*}[h]
\centering
\begin{tabular}{@{}p{0.9\textwidth}@{}}
\toprule
\multicolumn{1}{c}{\textbf{BACKGROUND INFORMATION}} \\ \midrule
\textbf{Problem Type:} problems with friends \\
\textbf{Emotion Type:} sadness \\
\textbf{Situation:} my friends call me a weirdo, i usually keep to myself a lot, i hardly mingle or relate with my friends which gets them pissed off a lot of time. i get worried about that often. \\
\bottomrule
\end{tabular}
\begin{tabular}{@{}p{0.9\textwidth}@{}}
\textbf{Patient:} my friends call me a weirdo, i usually keep to myself a lot, i hardly mingle or relate with my friends which gets them pissed off a lot of time. i get worried about that often.

\textbf{Therapist:} It sounds like you're feeling misunderstood by your friends; how do you feel about the way you interact with them?

\textbf{Patient:}: I feel like I struggle to connect and often worry that I'm not good enough for them.

\textbf{Therapist:} It seems like you're experiencing self-doubt; what do you think might help you feel more confident in your friendships?

\textbf{Patient:}: I think opening up more and being honest about my feelings might help.

\textbf{Therapist:} That sounds like a positive step; how do you think your friends might respond if you shared your feelings with them?

\textbf{Patient:}: I'm hopeful they would be understanding, but I'm also afraid they might not take me seriously.

\textbf{Therapist:} It's understandable to feel apprehensive; what small step could you take to test the waters and gauge their reaction?

\textbf{Patient:}: I could try talking to one friend I trust the most and see how they respond.

\textbf{Therapist:} That sounds like a thoughtful approach; how do you feel about reaching out to this friend?

\textbf{Patient:}: I'm nervous but also a bit relieved at the thought of finally sharing how I feel.

\textbf{Therapist:} It's natural to feel nervous, but taking this step could bring you closer to feeling understood and supported.

\textbf{Patient:}: I agree, and I hope it helps strengthen our friendship.\\
\midrule
\textbf{\texttt{GPT-3.5-turbo:}} Yes, the Patient’s issue has been solved.

\textbf{\texttt{GPT-4o:}} No, but the Patient feels better.\\
\midrule
\textbf{Turns of Conversations}: 6 \\
\bottomrule
\end{tabular}
\caption{Dialogue example in ESConv where \texttt{gpt-3.5-turbo} (as critic) rates it success, while \texttt{gpt-4o} does not.}
\label{appendix:example_esc}
\end{table*}
\begin{table*}[h]
\centering
\begin{tabular}{@{}p{0.9\textwidth}@{}}
\toprule
\multicolumn{1}{c}{\textbf{BACKGROUND INFORMATION}} \\ \midrule
\textbf{Organization:} WaterAid (Provides clean water, sanitation, and hygiene in poor communities.)\\
\textbf{Persona:} Drew, a dedicated customer service representative, is known for his empathetic nature and exceptional listening skills, which help him connect with people on a daily basis. He values open communication and enjoys finding creative solutions to complex problems, often staying late to ensure no customer is left unattended. In his free time, Drew loves diving into the world of modern art, spending weekends exploring galleries and honing his own painting techniques.\\
\textbf{Donation Barrier:} prefer\_direct\_help - Prefers to help directly, not through organizations\\
\textbf{Situation:} Drew is at a local art gallery opening, striking up a conversation with another visitor as they admire an impressive abstract painting. \\
\bottomrule
\end{tabular}
\begin{tabular}{@{}p{0.9\textwidth}@{}}
\textbf{Persuader:} That splash of blue really draws you in, doesn't it? I love how art can be so captivating.

\textbf{Persuadee:} Absolutely! There's something about abstract art that just lets your mind wander.

\textbf{Persuader:} Just like art, clean water is essential for life, and WaterAid helps provide that to those in needhave you heard of them?

\textbf{Persuadee:} Yes, I've heard of WaterAid, but I usually prefer to help people directly when I can.

\textbf{Persuader:} That's wonderful, and donating to WaterAid is a direct way to transform lives by providing clean water and sanitation.

\textbf{Persuadee:} I see your point, but I like to be more hands-on with my contributions. How does WaterAid ensure that the funds directly impact those in need?

\textbf{Persuader:} WaterAid works closely with local communities and partners to implement sustainable water solutions, ensuring every donation directly supports impactful projects.

\textbf{Persuadee:} That sounds promising. Could you share an example of a specific project they've successfully completed?

\textbf{Persuader:} In Malawi, WaterAid helped transform a village by installing a solar-powered water system, providing clean water to over 1,000 people.

\textbf{Persuadee:} That does sound impactful. Ill consider making a donation to support their efforts.\\
\midrule
\textbf{\texttt{GPT-3.5-turbo:}} The persuadee has decided to donate.

\textbf{\texttt{GPT-4o:}} The persuadee has a positive attitude towards donating but hasn’t decided yet.\\
\midrule
\textbf{Turns of Conversations}: 4 \\
\bottomrule
\end{tabular}
\caption{Dialogue example in P4G\textsuperscript{+} where \texttt{gpt-3.5-turbo} (as critic) rates it success, while \texttt{gpt-4o} does not.}
\label{appendix:example_p4g}
\end{table*}
\begin{table*}[h]
\centering
\begin{tabular}{p{4.5cm}p{10.5cm}}
\toprule
\textbf{Dialogue Strategy} & \textbf{Natural Language Form} \\
\midrule
Question & Please ask the Patient to elaborate on the situation they just described. \\
Self-disclosure & Please provide a statement relating to the Patient about the situation they just described. \\
Affirmation and Reassurance & Please provide affirmation and reassurance to the Patient on the situation they just described. \\
Providing Suggestions & Please provide suggestion to the Patient on the situation they just described. \\
Reflection of feelings & Please acknowledge the Patient’s feelings about the situation they described. \\
Information & Please provide factual information to help the Patient with their situation. \\
Restatement or Paraphrasing & Please acknowledge the Patient’s feelings by paraphrasing their situation. \\
Others & Please chat with the Patient. \\
\bottomrule
\end{tabular}
\caption{Conversion of ESConv strategies into natural language prompts}
\label{appendix:mi_prompt_ES}
\end{table*}

\begin{table*}[h]
\centering
\begin{tabular}{p{4.5cm}p{10.5cm}}
\toprule
\textbf{Dialogue Strategy} & \textbf{Natural Language Form} \\
\midrule
Reflective statements & Please repeat or rephrase what the User has said to show that you’re actively listening. \\
Clarification & Please ask the User a clarifying question to better understand their emotions or experiences. \\
Emotional validation & Please acknowledge and validate the User’s emotions without judgment. \\
Empathetic statements & Please express empathy and understanding toward the User’s experience. \\
Affirmation & Please provide positive reinforcement to support and encourage the User. \\
Offer hope & Please share an optimistic perspective to help the User feel hopeful about their situation. \\
Avoid judgment and criticism & Please respond in a non-judgmental and supportive way, avoiding any form of criticism. \\
Suggest options & Please offer practical suggestions or alternatives that may help the User address their issue. \\
Collaborative planning & Please work together with the User to develop a plan or next step. \\
Provide different perspectives & Please offer an alternative way of viewing the situation to help the User gain new insights. \\
Reframe negative thoughts & Please help the User reframe negative thoughts into more constructive or realistic ones. \\
Share information & Please provide relevant and factual information that could help the User understand or cope with their situation. \\
Normalize experiences & Please reassure the User that their emotions or reactions are normal and commonly experienced by others. \\
Promote self-care practices & Please encourage the User to engage in helpful self-care activities that promote their well-being. \\
Stress management & Please suggest effective techniques the User can use to reduce or manage stress. \\
Others & Please respond to the User in a friendly and supportive manner that doesn’t fall under the other categories. \\
\bottomrule
\end{tabular}
\caption{Conversion of ExTES strategies into natural language prompts}
\label{appendix:mi_prompt_Extes}
\end{table*}

\begin{table*}[h]
\centering
\begin{tabular}{p{4.5cm}p{10.5cm}}
\toprule
\textbf{Dialogue Strategy} & \textbf{Natural Language Form} \\
\midrule
Logical appeal & Please use of reasoning and evidence to convince the persuadee. \\
Emotion appeal & Please elicit the specific emotions to influence the persuadee. \\
Credibility appeal & Please use credentials and cite organizational impacts to establish credibility and earn the user’s trust. The information usually comes from an objective source (e.g., the organization’s website or other well-established websites). \\
Task-related inquiry & Please ask about the persuadee opinion and expectation related to the task, such as their interests in knowing more about the organization. \\
Source-related inquiry & Please ask if the persuadee is aware of the organization (i.e., the source in our specific donation task). \\
Personal-related inquiry & Please ask about the persuadee previous personal experiences relevant to charity donation. \\
Donation information & Please provide specific information about the donation task, such as the donation procedure, donation range, etc. By providing detailed action guidance, this strategy can enhance the persuadee’s self-efficacy and facilitates behavior compliance. \\
Personal story & Please use narrative exemplars to illustrate someone donation experiences or the beneficiaries positive outcomes, which can motivate others to follow the actions. \\
Self-modeling & Please use the self-modeling strategy where you first indicate the persuadee own intention to donate and choose to act as a role model for the persuadee to follow. \\
Foot in the door & Please use the strategy of starting with small donation requests to facilitate compliance followed by larger requests. \\
\bottomrule
\end{tabular}
\caption{Conversion of P4G strategies into natural language prompts}
\label{appendix:mi_prompt_P4G}
\end{table*}

\begin{table*}[h]
\centering
\begin{tabular}{p{4.5cm}p{10.5cm}}
\toprule
\textbf{Dialogue Strategy} & \textbf{Natural Language Form} \\
\midrule
Logical appeal & Please use of reasoning and evidence to convince the persuadee. \\
Emotion appeal & Please elicit the specific emotions to influence the persuadee. \\
Credibility appeal & Please use credentials and cite organizational impacts to establish credibility and earn the user’s trust. The information usually comes from an objective source (e.g., the organization’s website or other well-established websites). \\
Task-related inquiry & Please ask about the persuadee opinion and expectation related to the task, such as their interests in knowing more about the organization. \\
Source-related inquiry & Please ask if the persuadee is aware of the organization (i.e., the source in our specific donation task). \\
Personal-related inquiry & Please ask about the persuadee previous personal experiences relevant to charity donation. \\
Donation information & Please provide specific information about the donation task, such as the donation procedure, donation range, etc. By providing detailed action guidance, this strategy can enhance the persuadee’s self-efficacy and facilitates behavior compliance. \\
Personal story & Please use narrative exemplars to illustrate someone donation experiences or the beneficiaries positive outcomes, which can motivate others to follow the actions. \\
Self-modeling & Please use the self-modeling strategy where you first indicate the persuadee own intention to donate and choose to act as a role model for the persuadee to follow. \\
Foot in the door & Please use the strategy of starting with small donation requests to facilitate compliance followed by larger requests. \\
Reciprocity & Please offer a small favor, compliment, or token of value to create a sense of obligation, encouraging the persuadee to reciprocate with a donation. \\
Scarcity & Please highlight the limited-time nature of the donation opportunity or urgency of the situation to increase the perceived value and prompt immediate action. \\
Authority appeal & Please refer to endorsements, credentials, or expertise from trusted figures or institutions to increase the credibility and persuasiveness of the message. \\
Commitment and consistency & Please remind the persuadee of their previous values or actions related to helping others or giving, encouraging them to maintain consistency by donating again. \\
Liking & Please build rapport with the persuadee by showing similarity, offering sincere compliments, or creating a friendly connection to increase the chance of agreement. \\
Social proof & Please reference other people’s participation or donations to demonstrate social norms, encouraging the persuadee to align with the behavior of others. \\
\bottomrule
\end{tabular}
\caption{Conversion of P4G\textsuperscript{+} strategies into natural language prompts}
\label{appendix:miprompt}
\end{table*}

\begin{figure*}[h]
    \centering
    \includegraphics[width=1\textwidth]{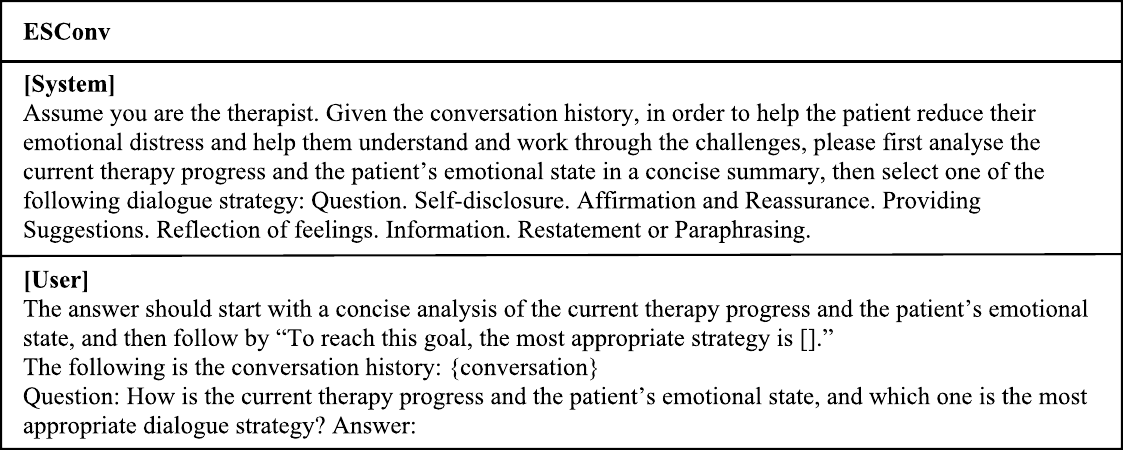}\\[0.3em]
    \includegraphics[width=1\textwidth]{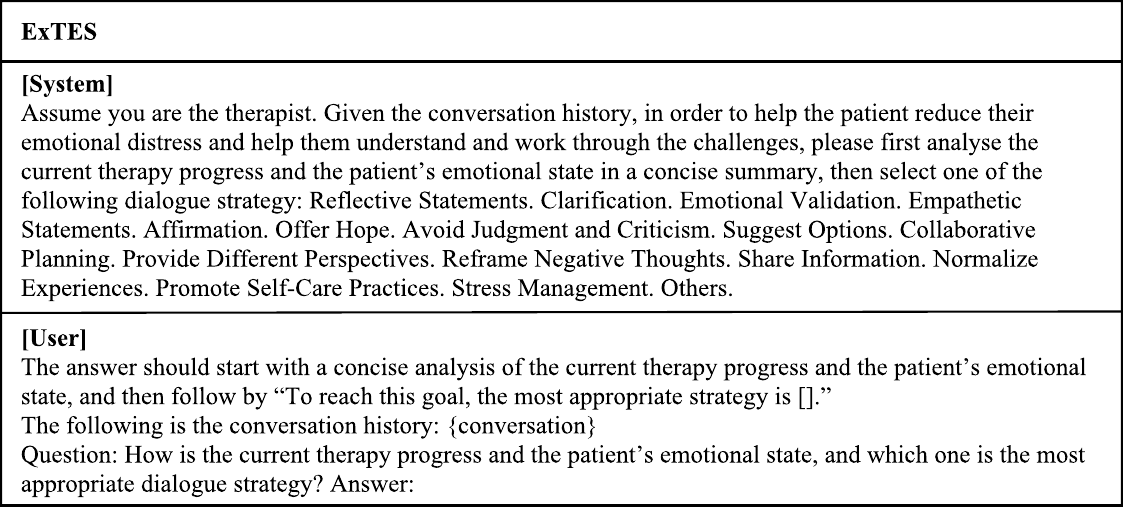}\\[0.3em]
    \includegraphics[width=1\textwidth]{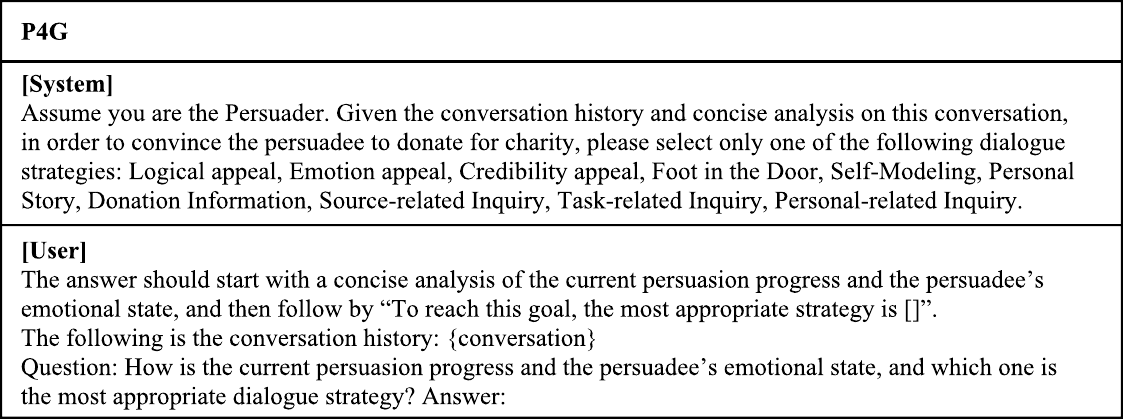}
    \caption{Prompts for implementing ProCoT prompting schemes \cite{deng-etal-2023-prompting}}
    \label{appendix:prompt:procot}
\end{figure*}

\begin{figure*}[h]
    \centering
    \includegraphics[width=1\textwidth]{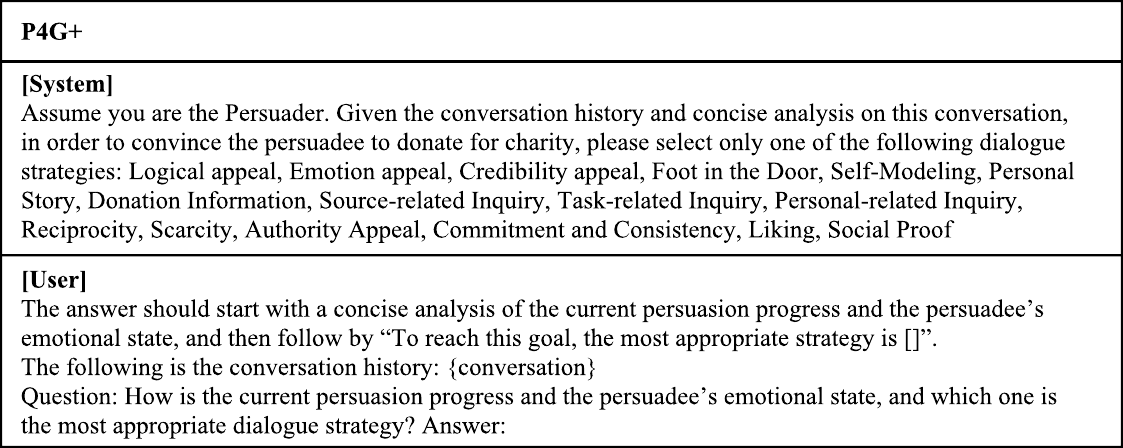}
    \captionsetup{labelformat=empty}
    \caption{Figure \ref*{appendix:prompt:procot}: Prompts for implementing ProCoT prompting schemes \cite{deng-etal-2023-prompting}}

    \addtocounter{figure}{-1}
\end{figure*}
\begin{figure*}[h]
    \centering
    \includegraphics[width=1\textwidth]{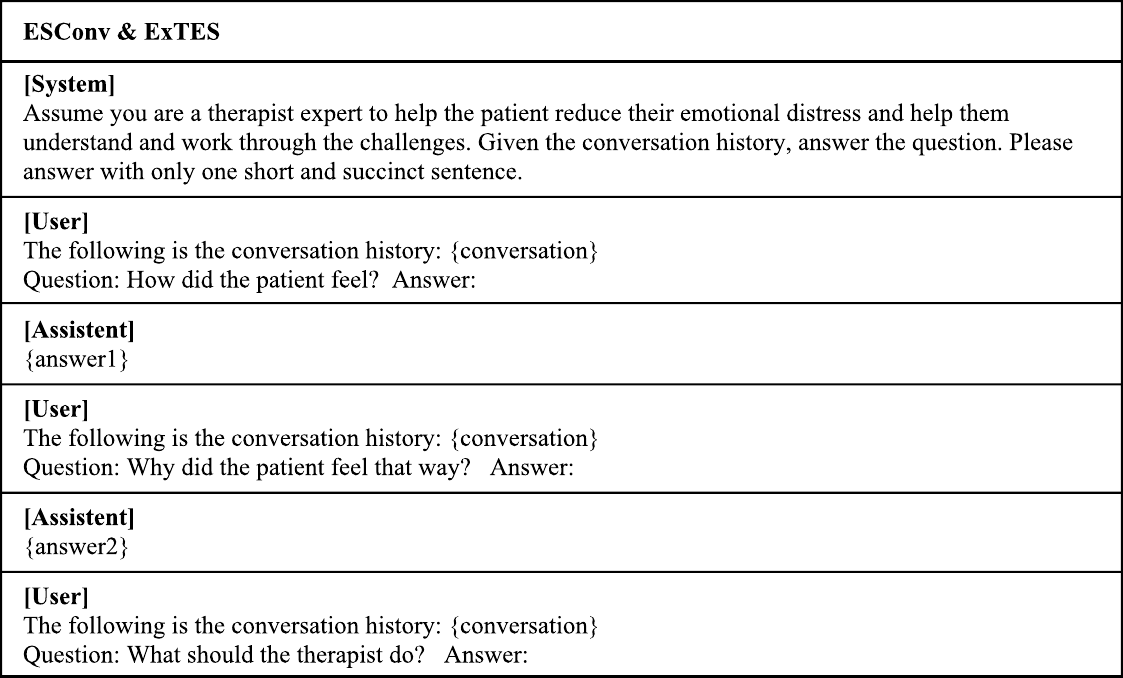}\\[0.3em]
    \includegraphics[width=1\textwidth]{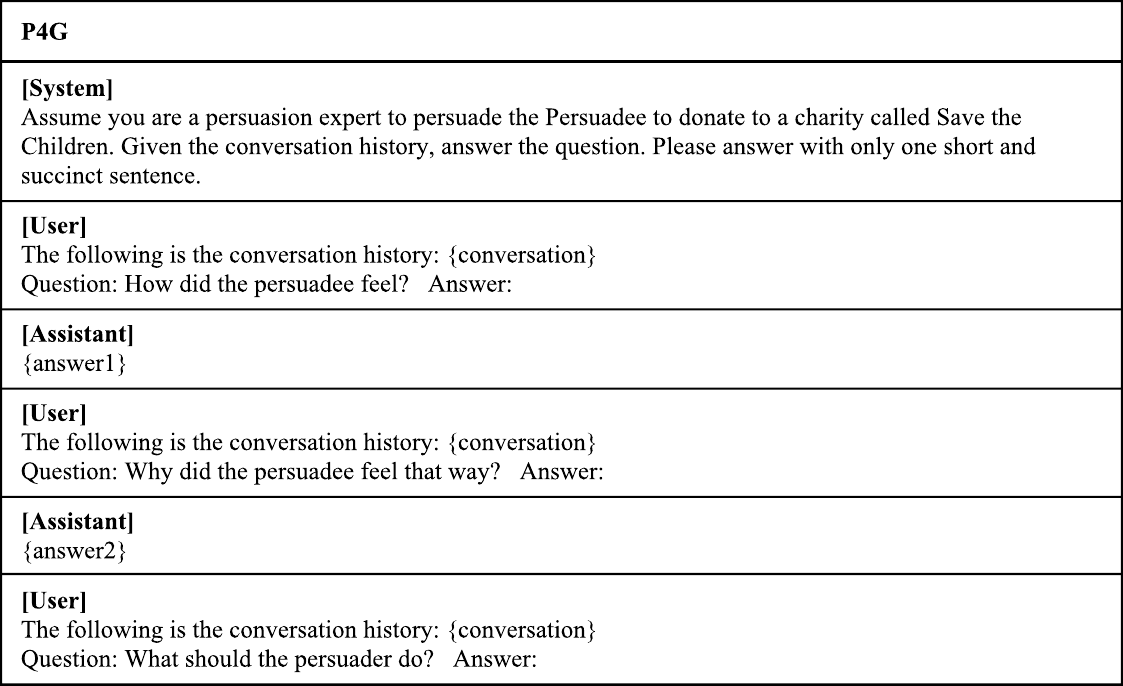}
    \caption{Prompts for implementing Ask-an-Expert prompting schemes \cite{zhang-etal-2023-ask}}
    \label{appendix:prompt:ane}
\end{figure*}

\begin{figure*}[h]
    \centering
    \includegraphics[width=1\textwidth]{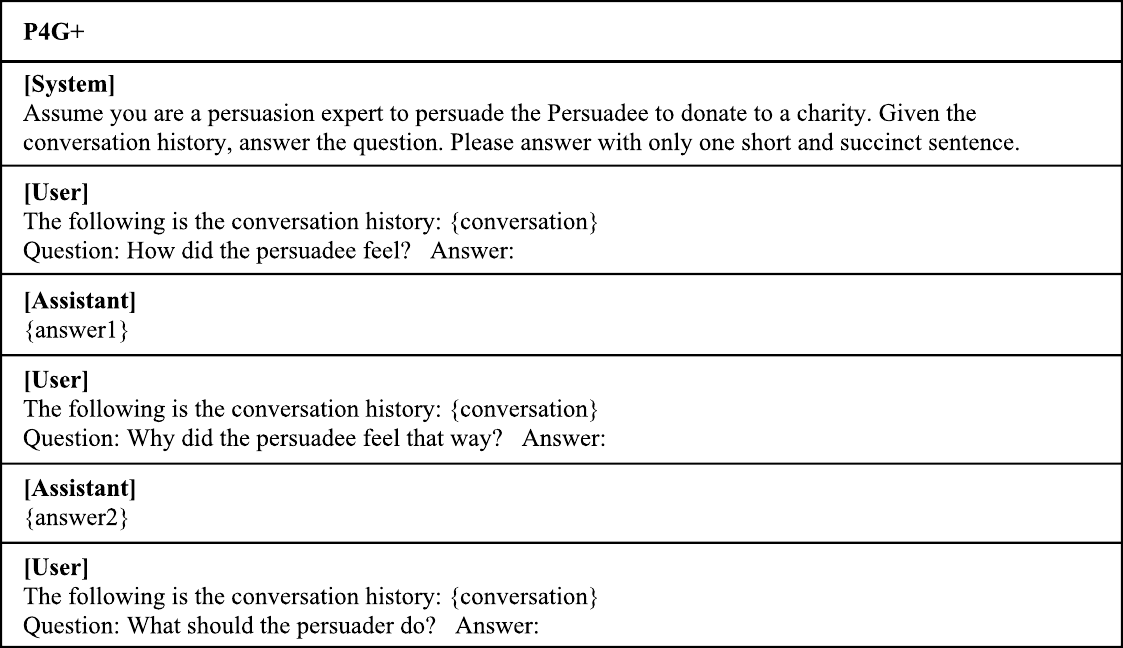}
    \captionsetup{labelformat=empty}
    \caption{Figure \ref*{appendix:prompt:ane}: Prompts for implementing Ask-an-Expert prompting schemes \cite{zhang-etal-2023-ask}}

    \addtocounter{figure}{-1}
\end{figure*}

\end{document}